\pdfoutput=1
\documentclass{article}

\usepackage[a4paper]{geometry}
\usepackage[T1]{fontenc}
\usepackage[utf8]{inputenc}
\usepackage[american]{babel}
\usepackage[autostyle=true]{csquotes}
\usepackage[american]{isodate}
\usepackage[autolanguage]{numprint}
\usepackage{expex}
\usepackage[linguistics]{forest}
\usepackage{booktabs}
\usepackage{adjustbox}
\usepackage{mathtools}
\usepackage{amssymb}

\usepackage{natbib}
\usepackage{hyperref}
\hypersetup{pdfauthor={Andy Lücking, Sebastian Brückner, Giuseppe Abrami, Tolga Uslu, Alexander Mehler}, pdftitle={Computational linguistic assessment of textbook and online learning media by means of threshold concepts in business education}, pdfsubject={Submitted to Frontiers in Education, July 2020}}

\newcommand{\gn}[1]{\textcolor{green!60!black}{\numprint{#1}}}
\newcommand{\rn}[1]{\textcolor{red!80!black}{\numprint{#1}}}

\begin{document}

\title{Computational linguistic assessment of textbook and online learning media by means of threshold concepts in business education}

\author{Andy Lücking\,$^{1,3}$ \and Sebastian Brückner\,$^{2}$ \and Giuseppe Abrami\,$^{1}$ \and Tolga Uslu\,$^{1}$ \and Alexander Mehler\,$^{1}$}

\date{\footnotesize$^{1}$Text Technology Lab, Institute of Computer Science, Faculty of Computer Science and Mathematics, \\ Goethe University Frankfurt, Frankfurt a.\,M., Germany \\
  $^{2}$Business Education, Johannes Gutenberg-Universität Mainz, Germany \\
  $^{3}$Laboratoire de Linguistique Formelle (LLF), \\ Laboratory of Excellence \enquote{Empirical Foundations of Linguistics} (EFL), Universit\'{e} de Paris, France}

\maketitle

\begin{abstract}

Threshold concepts are key terms in domain-based knowledge acquisition.
They are regarded as building blocks of the conceptual development of domain knowledge within particular learners. 
From a linguistic perspective, however, threshold concepts are instances of specialized vocabularies, exhibiting particular linguistic features.
Threshold concepts are typically used in specialized texts such as textbooks -- that is, within a \emph{formal} learning environment. 
However, they also occur in \emph{informal} learning environments like newspapers.
In this article, a first approach is taken to combine both lines into an overarching research program -- that is, to provide a computational linguistic assessment of different resources, including in particular online resources, by means of threshold concepts.
To this end, the distributive profiles of 63 threshold concepts from business education (which have been collected from threshold concept research) has been investigated in three kinds of (German) resources, namely textbooks, newspapers, and Wikipedia.
Wikipedia is (one of) the largest and most widely used online resources.
We looked at the threshold concepts' frequency distribution, their compound distribution, and their network structure within the three kind of resources. 
The two main findings can be summarized as follows: 
Firstly, the three kinds of resources can indeed be distinguished in terms of their threshold concepts' profiles.
Secondly, Wikipedia definitely appears to be a formal learning resource.
\end{abstract}


\section{Introduction: Threshold Concepts}
\label{sec:introduction}

In recent years, research on methods to facilitate the teaching, curriculum development and the diagnostic of competences acquired during higher education studies has intensified significantly in many disciplines, not only in Germany but also worldwide \citep{Nicola-Richmond:Pepin:Larkin:Taylor:2018,Zlatkin-Troitschanskaia:et:al:ED:2018}. Many tests have been developed that can measure the core competences in the respective domain both gradually and with structural validity \citep{Zlatkin-Troitschanskaia:et:al:ED:2019}. 
The core of these assessments as well as the design of the teaching-learning arrangements is always central focal content, which must be selected from curricula, textbooks and other learning media and adequately prepared for a target group, taking into account their specific learning requirements. 
Many factors have been identified in this respect, including prior knowledge, motivation and situational constraints (for general models of domain learning see e.g. \citealp{Ainsworth:2006,List:Alexander:2019,Goldman:et:al:2016}).
However, the content of the learning media itself has not yet been investigated on a large scale.
Due to advances in computational linguistics, the automatic processing of large corpora allows for just this kind of \enquote{text profiling} (c.f.\ \citealp{Mehler:Hemati:Welke:Konca:2020}).
In Sec.~\ref{sec:rationale-method}, the general linguistic and cognitive view on conceptual development in education is outlined.
Resting on the distinction between informal and formal learning media (on which see Subsec.~\ref{subsec:formal-informal}), in Sec.~\ref{sec:study} a study comparing (German) text resources (newspaper, Wikipedia and classical textbooks) of various kinds is reported and discussed. 
The results are connected to the general view in  Sec.~\ref{sec:general-discussion}.
In order to carry out the computational assessment on various resources, \emph{threshold concepts} are used.
But why are threshold concepts  particularly well suited for this task?

Due to limited teaching conditions, selection decisions require a consensus, since the content selected should, if possible, be such that it opens up the most comprehensive new understanding in a discipline and enables learners to solve multiple discipline-specific problems \citep{Davies:Mangan:2007,Meyer:Land:2006}. Instead of simplified content categories \citep{Kricks:Mittelstaedt:Liening:2013}, the threshold concepts approach has therefore been introduced into didactic discussions for several years \citep{Meyer:Land:2003}. 
%
The authors describe threshold concepts as \enquote{akin to a portal, opening up a new and previously inaccessible way of thinking about something} \citep[3]{Meyer:Land:2006}. Due to their special character within a discipline, they thus represent a threshold that needs to be crossed and that fundamentally changes the learner's understanding of the discipline. 
Concepts can thus describe regular processes, objects, theories, modeling methods on an abstract level, which contribute to the development of a comprehensive understanding of the learner within an individual discipline \citep{Sender:2017}.
In order to provide a more encompassing overview of threshold concepts, We emphasize some of their characteristic properties in the following %
Subsecs.~\ref{subsec:conceptual-change}--\ref{subsec:formal-informal}. 
In Sec.~\ref{sec:rationale-method} three perspectives on meaning are described (i.e. mental, referential and differential meaning) which are needed to embrace the research program's components. 
Referential meaning gives rise to expectancies concerning compound structures, differential meaning to text-bound threshold concept \enquote{webs} (see Subsec.~\ref{subsec:threshold-business-economics}), which are both operationalized in Sec.~\ref{sec:study}.

\subsection{Threshold concepts and conceptual change}
\label{subsec:conceptual-change}

The threshold concept approach can be distinguished from other approaches, which also imply content selection, on the basis of selected characteristics (e.g., core concepts, key concepts), since understanding the threshold concepts also changes the learner's perspective in the discipline as a whole and not only the single content \citep{Brueckner:Zlatkin-Troitschanskaia:2018}. \citet{Meyer:Land:2006} name five features that highlight the special characteristics of threshold concepts: 
\textit{transformativity}, \textit{irreversibility}, \textit{integrativity}, \textit{limitedness} and \textit{difficulty}. 
The sustained change of perspective on the discipline that the learners undergo, as described at the beginning, is attributed to the transformative characteristic. Not only the learned concept is re-evaluated, but also previously learned other concepts can be re-evaluated. Not only the cognitive, but also the affective and emotional disposition of the learners is addressed. 
Irreversibility refers to the remarkable circumstance that once a acquired, a threshold concept will not be forgotten easily, under normal conditions \citep{Meyer:Land:2005,Meyer:Land:2006}. 
The constant transfer and application of the acquired knowledge to a variety of known phenomena promotes the networking of knowledge. Integrativity thus leads to the fact that different knowledge structures, which previously could not be put into context for the learner, are increasingly brought into a semantic relation. Threshold concepts are also bounded, since the new conceptual spaces created by linking content-related ideas simultaneously create new boundaries that distinguish the discipline from other academic disciplines \citep{Meyer:Land:2005}.

\subsection{Threshold Concepts in Business and Economics}
\label{subsec:threshold-business-economics}

%
\textit{Opportunity costs} was the initial threshold concept that has been identified for the discipline of economics \citep{Meyer:Shanahan:2003} and has since been taken up in several studies \citep{Davies:Mangan:2007,Shanahan:Foster:Meyer:2006}. The critical discourse and empirical examination as to which concepts can be considered threshold concepts and which are important for the curriculum but not mandatory is ongoing and has since been discussed in a number of papers \citep{Davies:Mangan:2007,Lucas:Mladenovic:2009,Montiel:Gallo:Antolin-Lopez:2020}. 
The fact that the approach to a didactic design of learning environments in economics is currently attracting increasing research interest is also reflected in a number of recent articles \citep{Brueckner:Zlatkin-Troitschanskaia:2018,Hatt:2018,Lamb:Hsu:Lemanski:2019,Montiel:Gallo:Antolin-Lopez:2020,Sender:2017,vanMourik:Wilkin:2019}. Over the years, in addition to opportunity costs, a large number of concepts have been proposed and empirically tested in economics, e.g. on depreciation \citep{Lucas:Mladenovic:2009}, elasticity \citep{Reimann:Jackson:2006}, information asymmetry \citep{Hoadley:Tickle:Wood:Kyng:2015} and many more, on the basis of multiple research methods, e.g., using interviews with teachers, learners, videographies, curriculum analyses or standardized tests.
Some of the concepts require that a first encounter with a subject has already taken place and that the learner has a basic level of knowledge \citep{Davies:Mangan:2007}, for example, the concept of costs should be understood before the opportunity cost principle is understood.
The transition between thresholds is considered as conceptual change where \citet{Davies:Mangan:2007} distinguish three forms, that is, the \emph{basic}, \emph{discipline} and \emph{procedural} form of conceptual change. 
This three-part categorization has been taken up frequently, especially in recent years, by integrating further concepts from the economic sciences and further developing existing concept attributions \citep{Brueckner:Zlatkin-Troitschanskaia:2018,Hoadley:Tickle:Wood:Kyng:2015,Kricks:Mittelstaedt:Liening:2013,Lucas:Mladenovic:2009,Sender:2017,vanMourik:Wilkin:2019}.
%
%
Concepts documented along the basic threshold are accessible to most learners, as they are confronted with their everyday life (e.g. in their behavior as consumers) 
\citep{Davies:Mangan:2007}. 
At the level of the disciplinary threshold, the learner succeeds in developing and linking conceptual understandings based on a theoretically elaborated perspective, which are hardly accessible from everyday life. 
This concerns concepts that are mainly accessible within the economic sciences (e.g. the concept of opportunity costs, hedging; depreciation; see \citealp{Davies:Mangan:2007,Hoadley:Tickle:Wood:Kyng:2015,Lucas:Mladenovic:2009}). 
For this purpose the learner must have already developed some disciplinary understanding. The procedural threshold comprises concepts that are deeply integrated in the subject structures and require an understanding of modeling in economics. These are abstract modeling methods, procedures or argumentations that are used to analyze economic phenomena, but also to further develop economic theories (e.g. comparative statics, intertemporality; \citealp{Brueckner:Zlatkin-Troitschanskaia:2018,Davies:Mangan:2007,Sender:2017}).

\subsection{Threshold Concepts and troublesome language}
\label{subsec:troublesome-language}

The \enquote{troublesomeness} that learners experience in the transition of thresholds is reflected in particular in the fact that the disciplinary understanding of a concept differs from the individual understanding, the greater the effort of the learner is to understand the concept in a disciplinary adequate way \citep{Davies:Mangan:2007}. 
The difficulties students encounter with the understanding of disciplinary concepts and the frequency of corresponding usage situations may thus provide information about the learners' horizon of experience with these concepts.
For the learner to be able to compare his or her individual understanding of a concept with the disciplinary understanding and to initiate a  learning process, it is necessary that he or she can refer to previous (possibly naive) experiences.
Since the initiation of a learning process in a domain is in line with the generation of an understanding for the basic thresholds, it is necessary that the learners can connect their individual experience with a professional understanding of the concept. For disciplinary and modeling thresholds, however, learners already draw on basic domain-specific experience. \citet{Davies:Mangan:2007}  and \citet{Hoadley:Tickle:Wood:Kyng:2015}  show how the relations of a professional concept understanding can be illustrated in the form of a \enquote{web}.

\citet{Meyer:Land:2006} refer to the linguistic characteristics and contents associated with the concepts of a discipline that influence the crossing of thresholds as \enquote{troublesome language} and \enquote{troublesome knowledge}.
For example, compounds are related to conceptual problems with which learners are confronted \citep{Meyer:Land:2006}.
It is therefore important which concepts are introduced in learning materials and how this material is structured, how the concepts are networked within texts
\citep{Montiel:Gallo:Antolin-Lopez:2020}. 
From this brief outline we learn that the way in which threshold concepts manifest themselves in texts and how their meaning is related to each other as a result of these manifestations -- surface-structurally within the texts (text as product or artifact; \citealp{Dennett:1990}) and cognitively as the result of corresponding writing and reading processes (text as process; both individually and socially, \citealp{Clark:1992,Trueswell:Tanenhaus:2005}) -- has a major influence on the transition between the thresholds previously identified for disciplinary learning. Consequently, we will consider both manifestation regularities (in the sense of compounding) and networking regularities of threshold concepts (see Sec.~\ref{sec:study}).
Both strands are elaborated in Sec.~\ref{sec:rationale-method}: in order to locate threshold concepts within the larger research program of educational learning, it is necessary to distinguish (and to relate, of course) the cognitive states of learners, the meaning of expressions, and their textual (co-)occurrences. But at first we have to clarify and classify \emph{where} we expect to find threshold concepts.

\subsection{Threshold concepts, specialized vocabularies, and formal and informal learning}
\label{subsec:threshold-and-special-vocab}\label{subsec:formal-informal}

Since threshold concepts in business and economics are addressed by words it comes as no surprise that there is a connection to investigations from linguistics, in particular in studies of a certain  kind of a manner of speaking (a socio-, functo-, or technolect) known as \emph{specialized languages}, or the \enquote{language of science}.
A specialized language is more than just a specialized vocabulary since it involves grammatical aspects as well \citep[384]{Crystal:1997} -- however, the vocabulary is the most salient part of a scientific sociolect and  threshold concepts are no exception to this impression.
Accordingly, there is a branch of linguistics specialized on specialized languages \citep[see][for an introduction]{Roelcke:2010}, in particular in lexicography \citep{Fachsprachen:1998}.
%
%
Interestingly, lexicographic work on specialized vocabularies distinguishes three classes of scientific expressions: \enquote{technical terms, semi-technical terms and general vocabulary frequently used in a specialized domain} (\citealp[9]{Motos:2011}, quoted from \citealp[267]{Nagy:2014}).
%
Obviously, there is a coincidence with the three-fold distinction of threshold concepts into basic, discipline and procedural, which could be worth to pursue. 
The present study, however, investigates textual features with regard to threshold concepts, based on linguistic considerations concerning specialized languages.

%
Unlike in other industrial nations, \textit{business} or \textit{economics} has not yet been established as a school subject in Germany \citep{Schuhen:Kunde:2016}. 
%
The majority of first-year students at German universities usually have previous knowledge that was acquired in an \emph{informal context} \citep[cf.][]{Schumann:et:al:2010}.  
\emph{Formal learning} is therefore essentially institutionalized in schools, further education courses or universities and is thus explicitly initiated, accompanied, mostly qualified and certified and also perceived by the learners in a corresponding way \citep{Hofhues:2016}. 
Part of the institutionalization within schools is a curriculum, which includes a selection of teaching material. Institutionalized teaching materials are primarily textbooks.
Textbooks in analogue or digital form still enjoy the highest level of credibility compared to other educational media with regard to the quality of the information presented in them and are intensively researched in scientific institutions (see e.g. the Georg Eckert Institute for International Textbook Research at \url{http://www.gei.de/en/home.html}).  
%

Informal learning can -- taking into account the variety of definitions -- essentially be understood as learning \textit{en passant}; i.e. learning that takes place quasi implicitly when carrying out other activities (e.g., learn about costs when reading a newspaper article), is usually not consciously controlled by the learner \citep{Hofhues:2016,Neuweg:2000}. 
There is vast empirical evidence that the majority of first-year students in economics are rudimentarily familiar with economic concepts or have naive understanding -- only every second economic concept was understood correctly by approx. \numprint{7000} first-year students (students solved an average of 13 out of 25 subject tasks correctly) \citep{Schlax:et:al:2020}. 
The first-year students' knowledge of economics often comes from media that are not directly related to a learning-intended purpose (e.g. online magazines, news magazines, videos) \citep{Maurer:et:al:2019}, social interactions on financial topics (e.g. as a consumer in a supermarket or buying a mobile phone) \citep{Davies:Mangan:2007,Schuhen:Kunde:2016}, or other behavior with economic relevance (e.g. retirement planning). 
%

Traditionally, of course, the major resource for developing formal competences are textbooks \citep{Jadin::Zoeserl:2009,Maurer:et:al:2019}, whose didactic purposes include the introduction of special vocabulary, after all.
However, students not only consult textbooks -- their frequent digressions are due to the availability of online media. 
Wikipedia is often used to quickly obtain information on subject-specific concepts \citep{Jadin::Zoeserl:2009,Lim:2009,Maurer:et:al:2019}. 
According to the review of \cite[8]{Steffens:Schmitt:Assmann:2017}, Google and Wikipedia are the two topmost used internet services. 
That is, also non-university sources of information have to be taken into account \citep[as has been argued by][]{Maurer:et:al:2019}. 
However, Google is not an information source in itself; it is a search engine pointing at possible information sources.
Search requests from the domain of business and economics are likely to give results from newspapers, among others. 
It was also shown that students use digital media primarily for entertainment and communication, particularly in informal learning environments \citep{Blossfeld:et:al:2018,Steffens:Schmitt:Assmann:2017}. 
For these reason, the study described in Sec.~\ref{sec:study} looks at threshold concepts in textbooks, Wikipedia and newspapers, as representatives of different learning environments.

\section{Threshold concepts and linguistics: mental, referential and differential meaning}
\label{sec:rationale-method}

    
    

%

As outlined in Subsections \ref{subsec:conceptual-change} to \ref{subsec:troublesome-language}, threshold concepts from the disciplines of business and economics can and have to be approached from various perspectives: they are defined as specialized terms, they are building blocks of students' learning development and they are expressed by words.
Each of these perspectives corresponds to different scientific (sub-)disciplines (namely business and economics, learning psychology and education, and linguistics and lexicography, in that order; for a related view see \citealp{Lenci:2008}).
But how are they related?


\subsection{Different concepts of \enquote{threshold concepts}}
\label{subsec:methodology}

The outline in Section~\ref{sec:introduction} evinced a multifaceted role of threshold concepts: threshold concepts encode discipline-specific knowledge, they are important milestones in students' conceptual development, and they are realized as expressions of individual languages (like English or German). 
How can we make sense out of this?

According to a widely accepted sign-based conception, a word is a couple of a \emph{form} (hereafter also called \emph{expression}) and a \emph{meaning}.
The form side can be a token, an inflected morpho-syntactic expression of a type (lemma), or it can be the lemma itself.
%
%
With respect to the meaning side, any
scholar dealing with meaning faces a dilemma: she has to use meaningful words in order to describe the meaning of words \citep[cf.][]{Neurath:1932}.
In order to avoid vicious circles, a distinction between \emph{metalanguage} (the language used to describe meanings) and \emph{object language} (the language whose meanings are described) is to adhered to  (cf. Subsubsec.~\ref{subsubsec:concepts-meanings}).
The basic idea is that the metalanguage provides an interpreted descriptive framework according to which meanings (of the object language) can be specified.
In fact, there are (good) reasons to assume that such an approach cannot be circumvented -- the irreducibility of language principle (cf.\ either \citet{Wittgenstein:1984} for a usage-based view or \citet{Hjelmslev:1969} for a structuralist view of this argument). 

Now one can think that the meanings of words \emph{are} concepts. 
However, the concept a speaker associates with a word includes private episodes. 
Such private episodes do not belong to the shared (i.e. \emph{normative}) lexical meanings of words.
%
%
Accordingly, we also distinguish between the (idealized) lexical meaning of a threshold concept expression and (a student's) concept of it (Subsubsec.~\ref{subsubsec:indexed-concepts}).

But one can just look up the meaning of a word in a dictionary, can't one?
Although there is a kernel of truth in it, dictionaries completely avail themselves on the meanings of the object language of the dictionary; in other words, dictionaries contain \emph{paraphrases} of meanings (Subsubsec.~\ref{subsubsec:dictionary-concepts}).

\subsubsection{Lexical meanings}
\label{subsubsec:concepts-meanings}

The term \textit{meaning} applies to various relations.
Consider the list in (\nextx), where (\nextx a--c) are taken from \citep[30]{Murphy:2010}:
\pex 
\a \textit{Happiness} means \enquote*{the state of being happy}.
\a Happiness means never having to frown.
\a \textit{Glädje} means \textit{happiness} in Swedish.
\a By \textit{happiness} Peter means \textit{ecstasy}. 
\xe 

In (\lastx) only the first example (\lastx a) involves lexical meaning.
In (\lastx b) a consequence relation is expressed and in (\lastx c) a translation relation.
(\lastx d) finally is a about speaker meaning \citep{Linsky:1971}.
Speaker meaning is usually conceived as pragmatic while lexical meaning is semantic (\enquote{Speaker's Reference and Semantic Reference}, re-published in \citealp{Kripke:2011:RE}).

Besides lexical meaning there is \emph{compositional meaning} (which for instance accounts for the ambiguity within a simple sentence such as \textit{every dog chased a cat}, which as a relational (a single cat is chased) and a dependent (there are as many cats as dogs, that is, a plural interpretation of the singular noun phrase \textit{a cat}) reading; see e.g. \citealt{Zeevat:2018}). 

Lexical meaning has to be distinguished into \emph{sense} and \emph{denotation} (this distinction goes back to \citealp{Frege:1892:ORIG}).\footnote{This pair of kinds of meanings are often translated as \textit{sense} and \textit{reference}. However, since most semanticists would agree that reference is a pragmatic notion \citep{Searle:1969,Roberts:2019}, we reserve it for that purpose.}
The denotation relation gives rise to the phenomenon that natural language expressions \emph{are about} something in the first place.
The denotation of a word is the set of things \enquote{picked out} by that word.\footnote{The formal and logical properties of denotations are studied within model-theoretic semantics \citep{Zimmermann:2011:a}.} 
The sense of a word can be construed as the commonality of the things in its denotation \citep{Colung:Smith:2003}, where \enquote{commonality} includes rather loose family resemblance \citep{Wittgenstein:1984}, that is, a quality which to possess licenses to be part of that word's denotation \citep[cf.][26]{Murphy:2010}.
In other words, the sense, or the lexical meaning, connects word forms with external objects.
%
%
Obviously, only the sense but not the denotation of a word can be stored within a mental lexicon.
So, on this view meanings are both in the head (senses) and not in the head (denotations).
Accordingly, senses are composed out of \enquote{bits of thought} -- whatever it takes for a mental state to be a representation of the sense's quality.
In lexical semantics, senses are directly represented in terms of semantic components (see \citealp{Jackendoff:1983,Jackendoff:1991:b,Jackendoff:2002,Wierzbicka:1996,Pustejovsky:1995}).  
%
We know, however, of no lexical semantic analysis of threshold concept. Thus, describing the meaning of threshold concept expressions in terms of a (existing or specifically developed) metalanguage and their interactions wrt. to compositionality and inference could be a desideratum for further studies.

We have been slightly inconsistent so far: meanings have been ascribed to both words and thoughts.
The tension is resolved when considering that senses are \emph{types}, that is, abstract properties which have a normative (and therefore also coordinative) dimension (this issue will be briefly taken up in Subsubsec.~\ref{subsubsec:indexed-concepts}).
%
%
These sense types are tokened in thoughts of individuals. 
Accordingly, in cognitive sciences concepts are construed as  \enquote{temporary constructions in working memory} \citep[34]{Barsalou:1993}. 
Each speaker instantiating a lexical sense instantiates his or her \emph{perspective} or \emph{understanding} of the lexical sense, or indexed concept. 

\subsubsection{Indexed concepts}
\label{subsubsec:indexed-concepts}

A \textit{concept} is a psychological entity, namely  a mental representation and therefore a property of an individual. 
A concept in the sense of the threshold concept approach integrates a disciplinary perspective -- an normative  description of an economic fact or a principle identified by experts -- with the individual perspective -- the individual mental representations that the learner associates with a fact -- within learning, the individual perspective matches the disciplinary one \citep{Sender:2017}.
%
This means that 
\begin{itemize}
    \item concepts are not directly observable (they can be evinced by learning assessments or (neuro-)psychological testing, however);
    \item concepts are loaden with individual-specific content 
    (which partly accounts for invidual-specific understanding);
    \item that concepts are the place where learning takes place.
\end{itemize}

Now speakers have knowledge about the meaning of lexical items; that is, part of speakers' lexicalized concepts is \emph{their understanding} of the sense of an expression -- this is also one of the hallmarks of Cognitive Grammar \citep[29\,f.]{Langacker:2013}.\footnote{Despite claims that concepts and meanings are complementary contents \citep[e.g.][]{Barsalou:et:al:1993}. Note further that according to Cognitive Grammar \enquote{meanings are in the minds of the speakers who produce and understand the expressions} \citep[27]{Langacker:2013}. Obviously this claim can only be made because Cognitive Grammar lacks a notion of denotation, leaving it with the identity problem of conceptual content.}
Hence, the senses identified and modeled in lexical semantics are idealizations; these senses are only realized in meaning-making minds.\footnote{There are historical positions that postulate an objective existence of senses -- Frege's \citeyear{Frege:1892:ORIG} \enquote{third realm} is a classic example. However, in consideration of the overwhelming empirical evidence that mental content is bound to a working brain, there is no question any more that peoples' mind are the hosts of meanings (though not of meanings construed as denotations).} 
Given the necessary individual nature of concepts, we represent them as \emph{indexed mental states}, where the index refers to the concept-bearing individual.
For instance, $s13$'s (mnemonically for \enquote*{student with enrolment number 13}) concept of costs is  \enquote*{concept$_{s13}$(\textit{cost})}.
Note that since \enquote*{concept$_{s13}$(\textit{cost})} is indexed to $s13$, any element of it must be too, amounting to the fact that \enquote{sense(\textit{cost})} here represents $s13$'s understanding of the sense of \textit{cost} -- \emph{mutatis mutandis for any other index}. 
Thus, when we talk about \emph{the} meaning or \emph{the} concept of an expression, we rely on an idealization, namely the assumption that we share meanings and have a common understanding.
Of course, this issue has not gone unnoticed.
In fact, there are several genealogical reasons that prevent a \enquote{conceptual solipsisms}.
These include:
%
\emph{coordination} (\citealp{Lewis:1969}; meanings get coordinated between communities of language users \textit{via} situation of language use),  and 
%
\emph{evolution} (\citealp{Millikan:1984}; meanings  have a historic yet normative force acquired as biological functions in evolutionary processes).
%
%
Following a semiotic variant of the principle of methodological individualism \citep{Keller:1995}, socially accepted concepts have to be explained in terms of individual concepts (further examples are known from social ontologies; \citealp{Searle:2006}).
Following the advise of \citet[304]{Klein:Kracht:2014}, namely 
\enquote{the more we talk to each other, the easier it gets, and the more we can come to understand each other},  natural language dialogue is the best way for securing mutual understanding. 
Such an approach is actually pursued in learning studies, where, e.g., classroom interactions are observed. 
In particular nonverbal behavior of the learners provide evidence on their conceptualizations \citep{Cook:Goldin-Meadow:2006}, in line with the dictum that, for instance, manual gestures are \enquote{postcards from the mind} \citep{de:Ruiter:2007:a}.

\subsubsection{Dictionary concepts}
\label{subsubsec:dictionary-concepts}

While lexical semantics is a useful tool for linguistic analyses of word meanings (cf. Subsubsec.~\ref{subsubsec:concepts-meanings}), it is less useful for everyday use and computational applications.
After all, when one wants to know what a word means, one looks it up in a dictionary. 
According to the British English Online Dictionary\footnote{\url{https://dictionary.cambridge.org/}, accessed at \printdate{2020-05-14}.}, the meaning of \textit{cost} is \enquote{the amount of money that you need to buy or do something}.
In contrast to lexical semantics, a dictionary describes object language terms in terms of object language terms.\footnote{\citep[34]{Murphy:2010} is very explicit: \enquote{Such paraphrases, also called glosses, are indicated in single quotation marks. One must keep in mind, however, that these glosses are not themselves the meanings of the words (as they are represented in our minds) -- they are descriptions of the meanings of the words.}}
The sketch of meanings from Subsubsec.~\ref{subsubsec:concepts-meanings} suffices in order to make more precise what claim a dictionary entry makes.
\ex
sense(\textit{cost}) $\equiv$
\begin{forest}
[sense(NP)
  [sense(DET) [the]]
    [sense(N$'$) 
    [sense(N) [{amount of money}, roof]]
    [sense(S$'$) [{that you need to buy or do something}, roof]]
  ]
]
\end{forest}
\xe

The lexical meaning of \textit{cost} is the sense of the syntactic parse (compositional meaning) of the gloss. 
The reader learns the meaning of \textit{cost}, if he or she knows sense(NP).
%
%
%
Furthermore, in order to derive sense(NP) not only the lexical meanings but also the compositional meanings have to be computed. 
In order to avoid this, a further simplification can be made by abstracting away from compositional meanings.
Now the lexical meaning of \textit{cost} is related (but not equivalent any more) to the lexical meanings of the content words from the gloss, as in (\nextx)
\ex 
sense(\textit{cost}) is related to sense(\textit{amount}), sense(\textit{money}), sense(\textit{need}), sense(\textit{buy}), sense(\textit{do}), and sense(\textit{something})
\xe 

%

Interestingly, for the dictionary user (\lastx) is nearly as helpful as (\blastx). 
Most notably, however, dictionary concepts give rise to a notion of \emph{context} of a learning media \citep[cf.][]{Braun:Weiss:Seidel:2014}: the context in (\lastx) is just the collection of expressions of the dictionary gloss. But in general a context can be any stretch of text from a few words to entire corpora or online resources.
Given a context of expressions (dictionary entry, corpus, \ldots), the expressions are transferred into a claim about their senses, as is made precise in (\lastx).
What happens here is that a statement about meanings is given in purely relational manner in terms of the object language -- just like in a dictionary paraphrase.
That is, (\lastx) exemplifies the scheme of a \emph{differential} rather than \emph{referential} approach to word meaning \citep{Sahlgren:2008}.\footnote{This line of thought is rooted in structuralism \citep{de:Saussure:1916:ORIG,Hjelmslev:1961}.}
Ultimately based on word frequency measures within text corpora, the relata of an expression can also be assigned different strengths by means of vector-valued word representations \citep{Spaerk-Jones:1972,Mikolov:et:al:2013,Levy:Goldberg:Dagan:2015} -- reflecting their respective \enquote{importance}.
So in a computational way, a dictionary entry can be conceived as a collection of expressions.\footnote{Dictionary approaches are developed into directions that make use of an extended notion of context \citep[multimodal networks,][]{Bruni:Tran:Baroni:2014} and try to deal with compositionality  \citep[see the discussion in][]{Boleda:Herbelot:2016}. So the term \enquote{dictionary} becomes a bit to narrow for these developments.}
Hence for \textit{cost}, in addition to sense(\textit{cost}) and concept$_x$(\textit{cost}) there is further concept, dict(\textit{cost}), the  dictionary concept of the expression.
Now dictionary concepts have a further property which is useful for present purposes: for any two non-identical contexts $c_1$ and $c_2$, the dictionary concept of a random expression will differ with respect to $c_1$ and $c_2$. In other words, dictionary concepts are  \emph{text-bound}, and text-boundedness is a prerequisite for comparing different resources in the first place.
From a learning perspective, an interpreter of a dictionary entry has to entertain an indexed concept for each of its elements -- amounting to the transient nature of threshold concepts and the mental linkage emphasized in Subsec.~\ref{subsec:threshold-business-economics}. 
In sum, with sense($\cdot$) we have a cognitive but not text-bound notion of meaning at our disposal, and with dict($\cdot$) a text-bound but not cognitive one. Let us finally use these notions in order to spell out linguistically driven expectancies with respect to the use of threshold concepts within formal and informal text corpora.


\subsubsection{Concept expressions and the \enquote{Law of Denotation}}
\label{subsubsec:concept-expressions}

Since the expressions of a word is its only observable one, how can one make use of dictionary concepts?
(Lexical) semantics discovered a couple of principles which are productive in this respect.
The most important one for current purposes is 
what \citet[36]{Murphy:2010} calls the \emph{Law of Denotation} (LoD): the \enquote{bigger} a word's sense (i.e. the more conditions that it places on what counts
as a referent for that word), the smaller its extension will be.
There are several phenomena to which this principle applies.
For instance, the hypernym--hyponym relation fulfills the law of denotation, as does compounding. 
A broader term like \textit{dog} has less lexical meaning components than a narrower term like \textit{dachshund}.\footnote{In this case one must of course know that \textit{dachshund} is a hyponym of the hypernym \textit{dog}. According to dictionary approaches, such knowledge is part of the speaker's mental lexicon, according to conceptual semantics it is computed based on semantic componential representations.\label{fn:lex-rels}}
Since the modifying noun of a nominal compound adds its meaning in some way or other to the head noun, the law of denotation is trivially fulfilled. 

Since every expression is bound up with a sense,\footnote{This is less clear, however, for syncategorematic expressions such as conjuncts. However, since they do not \emph{remove} any sense components, they do no harm to the generalization.} larger constituents are necessarily accumulative (in fact, compositional).
Now assuming expressions, sentences or discourses to be coherent (a notion on which see \citet[21, and various other places]{Asher:Lascarides:2003} and \citet[208 f.]{Ginzburg:2012}), this gives rise to the simple but useful generalization: \emph{the more expressions, the more elaborate the combined sense} (where \enquote{combined} is intended to cover both compositional derivation as well as accumulation).

The relation between senses and denotations is regimented by LoD.
It applies likewise to words, phrases and sentences. 
The more fine-grained the senses of these constituents, the more detailed are their denotations. 
The connection to sciences and the language of sciences is obvious: (natural) sciences aim at precise descriptions of the world. 
That is, scientific languages are about very detailed denotations. 
In order to achieve this level of detail, guided by LoD, the expressions of the specialized vocabularies need to have elaborate senses, which, by dint of compositional meanings, gets even more specific in phrases and sentences. 
Since natural languages are devices of ontology construction, as has been pointed out by some versions of semantics  \citep[e.g.][]{Barwise:Perry:1983}, it is also possible to \enquote{postulate new denotations}, so to speak, as has famously been done in the history of physics several times, for instance.
LoD and \textit{making things precise} has repercussions to linguistic expressions. 
Against this backdrop, we discuss observable features of expressions of threshold concepts in the following.

\subsection{Linguistic features}
\label{subsec:ling-features}

Following the guideline that threshold concepts are instances of specialized vocabularies, we expect their expressions 
to exhibit the following features:
\begin{itemize}

\item  \textbf{compounding potential}. Of how many compounds is an expression a part? 
The compounding potential is a long-known feature of specialized vocabulary where specialized languages are characterized by a large number of compounds  \citep{Widdowson:1974}.
It has also been highlighted by business and economics studies on threshold concepts (cf. Subsection~\ref{subsec:troublesome-language}; \citealp{Meyer:Land:2006}).
In light of the above-mentioned specificity demand of languages of science, this feature is expected.
But why are compounds semantically specific and distinguish themselves from \textit{prima vista} synonymous syntactic realizations?
Most nominal compounds (that are compounds whose head is a noun while the modifying component may be an adjective (\textit{green tea}), a verb (\textit{swimming pool}), or a further noun (\textit{football})) are determinative, meaning that the modifying expression determines the head noun.
For instance, a football is not just a ball, but a ball meant to be moved along by one's feet.
%
%
But there are more interesting properties of compounds.
Most importantly, a compound induces a \emph{kind reading} \citep{Buecking:2010}. 
Given this feature, we expect compounding (as a form of name-giving) to be coupled to the dynamic ontological modifications within the sciences, as is evinced by findings for specialized vocabulary  \citep{Widdowson:1974}.

If we conceive the kind-reading of compounds in relation to LoD and the specificity demands of scientific languages, a few trends can be derived:
\begin{enumerate}
    \item For all compounds that share the same threshold concept expression head it holds that the more modifying constituents the compound has, the more specific it is.
    This follows trivially from sense accumulation.
    For instance, both \textit{Grenzkosten} \enquote*{terminal cost} and Marginalkosten \enquote*{marginal cost} are more specific than Kosten \enquote*{cost}.
    
    \item The inverse formulation of the previous item is that the more specific a given threshold concept head is, the less compounds it will show. 
    Note that this is a recursive notion: (more) complex compounds may consist of (less) complex heads. 
    
    \item Going from expressions to the use of these expression in sentences and texts it is very likely that the more compounds a sentence or text contains, the more specific the sentence or text is (see also the following linguistic feature, \enquote{large nominal groups}).
\end{enumerate}

These trends can directly be read off the concept expressions.

\item \textbf{large nominal groups}. Related to the compounding potential is the elaborateness of the whole nominal group of which a concept expression (compound or not) is a part.
Expressions of specialized vocabularies tend to occur in elaborate environments \citep{Strevens:1977}.
Contexts of elaborateness are constructed by adjectives and relative clauses (mainly restrictive ones).
Obviously, nominal groups are more specific according to LoD.
This features is a further linguistic feature of threshold concept expressions to look. 

\item  \textbf{\enquote{web of threshold expressions}}. Based on postulations of threshold concept research from Subsec.~\ref{subsec:threshold-business-economics}, concept expressions are to be expected to be related to each other -- that is, forming a \enquote{web} of threshold expressions \citep{Davies:Mangan:2007}. 
Thus, in terms of Subsubsec.~\ref{subsubsec:concept-expressions} we can make the claim more precise in saying that the web of threshold concepts is a context of weighted expressions where  the context consists exclusively of threshold concepts.
Now the different contexts under consideration (textbooks, newspaper, Wikipedia) trivially give rise to different dictionary concepts.
However, since the different contexts are an independent variable, differences can point at meaningful differences in the independent variable (i.e. contexts).
Further support for this claim comes from qualitative investigations of specialized vocabularies, where the context is accredited to be most important feature of special terms \citep{Vankova:2018}.
From that we can derive the expectation that the web of threshold concepts is \enquote{stronger woven} in formal than in informal contexts.

\end{itemize}

From Subsecs.~\ref{subsec:formal-informal} and  \ref{subsec:threshold-and-special-vocab} we take the further assumption that resources from formal learning environments are more specific than resources from informal learning environments.
Now the, the following biconditional working hypotheses can be derived:\footnote{Here we focus on threshold concepts within formal and informal learning contexts. For an assessment of the three classes of threshold concepts -- basic, discipline, modeling -- see the study of  \citet{Brueckner:Luecking:2019:b}.}
\begin{itemize}
    \item \textbf{WH1:} Formal corpora show \enquote{longer} compounds than informal ones, that is, for a given threshold expression head, formal corpora have more modifying constituents.
    
    \item \textbf{WH2:} There are more compounds involving threshold concepts (regardless of being heads or not) within formal corpora than in informal corpora.
    
    \item \textbf{WH3:} The threshold concepts within formal corpora are part of larger nominal groups than in informal corpora.
    
    \item \textbf{WH4:} The \enquote{web of threshold concepts} derived from formal corpora gives rise to a stronger connected threshold concept context than the one derived from informal corpora.
\end{itemize}

In the following, some of the general working hypotheses are operationalized into network-theoretical hypotheses.
To this, 63 threshold concepts (see Appendix A) are compared across several corpora where the textbook corpus consists of the textbooks listed in Appendix B.

\section{Study}
\label{sec:study}

\subsection{A Two-part Procedure for Measuring the Use of Threshold Concepts}
\label{sec:A Two-part Procedure for Measuring the Use of Threshold Concepts}

To test Working Hypothesis WH2, we develop a two-part procedure to measure significant differences in the use of threshold concepts.
Our first aim is to quantify the difference in the specificity of uses of threshold concepts. 
In order to operationalize this notion, we start from the following assumptions:
\begin{itemize}
\item 
%
The more often a threshold concept $x$ manifests itself as a component in compounds and the higher the frequencies of these compounds in corpus $C$, the higher the \textit{degree of specification} of $x$ and thus its use in $C$. 
We call this sort of specificity \textit{compounding-related specificity} or just \textit{c-specificity} of $x$ in $C$.
Furthermore, the more frequently the concept occurs in $C$ as a whole, the higher its polytextuality in the sense of \citet{Koehler:1986} (i.e.\ the higher the number of sentences by which it is semantically specified), the higher its degree of specification. 
We call this sort of specificity \textit{sentence-related specificity} or just \textit{s-specificity}.
And the higher the number of threshold concepts with the higher degrees of c- or s-specificity, the higher the overall specificity of this set of concepts in the underlying corpus. 
\item The more c- or s-specific the use of a threshold concept in a corpus, the more detailed and differentiated knowledge can be acquired about this concept by reading texts of this corpus (i.e., the larger the context of the dictionary concept of the threshold concept expression in question).
\end{itemize}

Starting from these considerations we arrive at the following hypothesis about the difference between formal and informal language corpora (manifesting formal and informal learning contexts) in terms of the c- and s-specificity with which they manifest threshold concepts: 
\begin{itemize}
\item[] \textbf{H1a:} \textit{The use of threshold concepts in formal language corpora is more c- or s-specific than in informal language corpora.}
\end{itemize}

%
Our second aim is to quantify the differences in the associative networks of threshold concepts as induced by corpora of three different genres, that is, of press communication, encyclopedic communication and technical communication. 
From Subsec.~\ref{subsec:formal-informal} we know that newspapers are an example for informal learning contexts, whereas textbooks make up formal contexts. Since to our knowledge there is no \textit{linguistic} judgment of Wikipedia in this respect yet, we remain neutral and will see how Wikipedia compares to formal and informal resources used in the following.
For this purpose, we start from the following consideration:
\begin{itemize}
\item The greater the differences in the ways threshold concepts are used in two corpora, the more different the associative relations that can be learned as a result of reading homogeneous subsets of texts of these corpora.
\end{itemize}
By a \textit{homogeneous subset} we mean a set of texts sampled from the same corpus.
It should be noted that we do not directly observe the acquisition 
of semantic associations between threshold concepts. 
Rather, this acquisition will be be estimated by means of word embeddings \citep{Mikolov:et:al:2013}.
The embeddings are compared for the purpose of measuring the semantic associations of the embedded concepts, in the sense of the \textit{Weak Contextual Hypothesis} (WCH) of \citet{Miller:Charles:1991}:
words that tend to be used in similar contexts are then regarded as semantically similar and correspondingly more strongly associated. 
If a corpus exhibits such contextual similarities, reading subsets of texts from that corpus makes the acquisition of corresponding syntagmatic or paradigmatic associations, as we assume, more likely. 
Thus, if the semantic associations of a corpus deviate significantly from those that can be expected, for example, from a thematically similar corpus of textbooks, this may have negative consequences for the acquisition of the concepts concerned. 
Even if we do not investigate this consequence ourselves, we at least measure the previously mentioned similarity or dissimilarity of association networks.
These considerations are a prerequisite for operationalizing the falsification of the following hypothesis about the difference between formal and informal language corpora in terms of the semantic networking of threshold concepts:
\begin{itemize}
\item[] \textbf{H1b:} \textit{Due to their usage contexts in formal language corpora, threshold concepts are more strongly associated than due to their usage in informal language corpora.}
\end{itemize}

By falsifying the alternative hypotheses of H1a and H1b, we obtain evidence that the threshold concepts we are looking at are used significantly differently in the genres under consideration, insofar as their uses correspond to different degrees of specificity (a), while spanning different semantic networks (b).
However, what differs in two ways, in that it induces the acquisition of concepts of different specificity (\textit{node-related}) and different associations (\textit{edge-related}), ultimately represents a different learning basis or learning context.
From this point of view, it becomes clear that we understand the structure induced by threshold concepts as a \textit{network of concept nodes and their association relations}, whose \enquote{shape} depends on what is said about them in the underlying corpus or how they are specified by means of compounding.
More precisely, let $T= \{a_1, \ldots, a_n\}$ be a set of threshold concepts and $C= \{x_1, \ldots, x_m\}$ a text corpus. 
Then, we denote by 
\begin{equation}
C(T) = (V, E, \mu, \nu, \lambda)    
\end{equation}
the \textit{Threshold Concept Network} (TCN) induced by $C$ over $T$ where $E\subseteq V^2$, $\mu\!: V\to \mathbb{R}_0^+$ is a function measuring the specificity $\mu(v)$ of each $v\in V\subseteq T$ in $C$, $\nu\!: E\to \mathbb{R}$ is a function measuring the semantic association $\nu(\{v,w\})$ between $v$ and $w$ for each $\{v,w\}\in  E$ and $\lambda\!: V\to T$ is an injective vertex labeling function.
More specifically, $\nu\{v,w\}$ is the cosine similarity of the embedding vectors computed for $v$ and $w$, respectively, by the operative embedding method by exploring $C$.

Let $C_i(T) = (V_i, E_i, \mu_i, \nu_i, \lambda_i)$ and $C_j(T) = (V_j, E_j, \mu_j, \nu_j, \lambda_j)$ be two TCNs induced by the corpora $C_i$ and $C_j$.
For any pair of vertices $v\in V_i, w\in V_j$, for which $\lambda_i(v) = \lambda_j(w)$, we will write $\dot{v} = \dot{w}$.
%
%
To operationalize the falsification of H1a and H1b, we now specify the functions $\mu$ and $\nu$ in more detail:
%
\begin{itemize}
\item \textit{On $\mu$ and H1a:}
We consider a simple frequency-related definition of $\mu$, according to which $\mu(v)$ corresponds to the number of tokens of the lemma $v$ in $C$ plus the number of occurrences of compounds in $C$ that contain $v$ as a component (\textit{c- $+$ s-specificity}). 
A first variant of $\mu$, denoted by $\mu'$, considers only the former number (\textit{c-specificity}), a second, denoted by $\mu''$, only the latter number (\textit{s-specificity}).
Let $\mu$ be any of these variants, then we derive the following rank-frequency distribution
\begin{eqnarray}
{\mu}(V) = ((v_{i_1},\mu(v_{i_1})), \ldots, (v_{i_n},\mu(v_{i_n})),\; \mu(v_{i_1}) \ge \ldots \ge \mu(v_{i_n}),\; v_{i_1},\ldots, v_{i_n} \in V
\end{eqnarray}
for which we compute the exponent $\alpha$ of the power law that best fits this rank distribution. 
In this way, we test the skewness of the distribution of the specificities of threshold concepts as induced by $C$: 
the higher the value of $\alpha$, the faster the frequency-related transition from high-rank (frequent or highly specified) to low-rank (rare or rarely specified) concepts; note that we always consider  small numbers of concepts for the distributions, so the slope cannot be the result of a larger number of rare concepts and especially \textit{hapax legomena}.
The alternative to H1a is now considered falsified if the corpus length-normalized rank specificity distribution of formal language corpora is above that of informal language ones, under the condition of a Zipfian, power law-like character of such distributions as normally observed for word frequency distributions \citep{Zipf:1949,Tuldava:1998} and also assumed for threshold concepts.
Beyond that, we assume that power laws better fit the use of threshold concepts in textbook corpora or in formal language corpora in general than in informal language corpora (e.g.\ of press communication).
Furthermore, we assume that the rank specificity distributions of formal language corpora differ significantly from those obtained for informal language corpora.
Finally, we assume that the rank correlation between the rank specificity distributions of formal and informal language corpora is lower than in cases where the corpora manifest either both formal or informal language -- provided that these corpora are all sufficiently similar thematically.
If we succeed in falsifying the alternative to H1a in these senses, we get the information that formal language contributes to the development of more specific threshold concepts, the specificity distribution of which follows a Zipfian distribution in a more pronounced and significantly different way compared to corpora of informal language, that the specificity of the concepts in the latter corpora tends to be lower, and that, finally, thematically and formally similar corpora are more similar to each other than corpora of different formality.
%
%
\item \textit{On $\nu$ and H1b:} 
The association strength of TCNs in relation to the degree of formality of the underlying corpus will be measured using methods of network theory \citep{Newman:2010:a} and especially of the theory of linguistic networks \citep{Mehler:Gleim:Gaitsch:Uslu:Hemati:2020}.
More specifically, we test H1b by quantifying the densities of TCNs derived from different corpora using the approach of \citet{Mehler:Hemati:Welke:Konca:2020}.
That is, we utilize the notion of $\alpha$-cuts, as introduced in the description of fuzzy sets, and apply it to weighted graphs as follows:
let $C(T) = (V, E, \mu, \nu, \lambda)$ be a TCN. 
Then we define:
\begin{alignat}{2}
a(C(T)) &= (\alpha_{1}, \ldots, \alpha_{l})^T 
\\ 
\alpha_{1} &= \min \{ s(\nu(e))\mid e\in E \}\\
\forall k\in \{2,\ldots, l\}\!: \alpha_{k} &= \min \{ s(\nu(e)) \mid s(\nu(e)) > \alpha_{k-1} \}\\
\forall e\in E\!: s(\nu(e)) &= \frac{\nu(e) - \text{Min}}{\text{Max} - \text{Min}}\in [0,1]
\end{alignat}
where $\text{Max}$ ($\text{Min}$) is the theoretical maximum (minimum) that $\nu$ can assume. 
Then we define the $\alpha$-cut of $C(T) = (V, E, \mu, \nu, \lambda)$, that is, the so-called \textit{alpha-cut graph} $C(T,\alpha) = (V, E|_\alpha, \mu|_\alpha, \nu|_\alpha, \lambda|_\alpha)$ where
\begin{alignat}{2}
E|_\alpha &= \{ e\in E \mid \nu(e) \ge \alpha \}
\end{alignat}
and $\mu|_\alpha, \lambda|_\alpha$ are the restrictions of $\mu, \lambda$ to the vertex set induced by $E|_\alpha$ and where $\nu|_\alpha\!: E|_\alpha\to [0,1], \forall e\in E|_\alpha\!: \nu|_\alpha(e) = s(\nu(e))$.
This allows us to define the graph series
\begin{alignat}{2}
\mathit{cuts}(a(C(T))) &= (C(T,\alpha_1), \ldots, C(T,\alpha_l))
\end{alignat}
Finally, for any graph index $\iota\!: \mathbb{G}\to \mathbb{R}$, we get an series of index values:
\begin{alignat}{2}
\iota(\mathit{cuts}(a(C(T)))) &= (\iota(C(T,\alpha_1)), \ldots, \iota(C(T,\alpha_l)))
\end{alignat}
In this paper, we experiment with graph cohesion and graph clustering \citep{Newman:2010:a}.
In the case of any such graph index we want to know (1) how early, (2) how fast, and (3) how different from the TCN series based on the comparison corpora, the series of alpha-cut graphs calculated for a given corpus either decays or increases.
Now Hypothesis H1b is considered falsified if the cohesion of the series of alpha section graphs calculated for the textbook corpus decreases later than in non-textbook corpora, and in such a way that the behaviors of these series differ significantly from each other.
Further, we expect the same behavior with regard to the corresponding series of transitivity values.
\end{itemize}

In a nutshell: 
H1 is considered falsified if the alternative hypotheses to H1a and H1b are falsified.
If such a double falsification succeeds, we obtain evidence that formal language corpora support the development of more strongly specified threshold concepts that are at the same time more strongly associated with each other or semantically networked. 
According to our guiding idea, such an observation is linked to the assumption that reading formal language corpora facilitates the acquisition of threshold concepts according to the associated learning objective.

\subsection{Data \& Preprocessing}

\begin{table}[htb]
\centering
\begin{tabular}{ lrrc } 
\toprule
 & No. of Articles & No. of Token & Period of publication  \\ 
\midrule 
SZ-Eco & \numprint{288792} & \numprint{85826410} & 1992--2014 \\ 
SZ-All  & \numprint{1707666} & \numprint{630588082} & 1992--2014\\ 
WP-Eco  & \numprint{653397} & \numprint{265063077} &  2001--2016 \\ 
WP-Top-1  & \numprint{37895} & \numprint{20090166} & 2001--2016 \\ 
WP-Top-3  & \numprint{71013} & \numprint{28145793} & 2001--2016 \\ 
WP-All  & \numprint{1760875} & \numprint{736071291} & 2001--2016 \\ 
ZEIT  & \numprint{184186} & \numprint{179327441} & 1994--2014\\ 
TB  & \numprint{14} books & \numprint{2326374} & 2015--2020 \\ 
\bottomrule
\end{tabular}
\caption{Summary of corpora used in the study. See main text for a description.}
\label{tab:corpora}
\end{table}

We consider corpora from press communication, encyclopedic communication and technical communication (see Tab.~\ref{tab:corpora}):
\begin{enumerate}
\item \textit{Corpus SZ-Eco:} 
as an informal language corpus of texts about economics, we process \numprint{288792} texts from the \textit{Süddeutsche Zeitung} (SZ) all of which belong to the register \textit{Wirtschaft} (\textit{economics}) -- see Table \ref{tab:corpora} for the corpus statistics.

\item \textit{Corpus SZ-All:}
SZ-Eco is contrasted with SZ-All, that is, the corpus of al \numprint{1707666} articles of SZ published in the years 1992 to 2014 (see Table \ref{tab:corpora}).
In this way we get access to the usage regularities of threshold concepts in arbitrary press articles of whatever topic.

\item \textit{Corpus WP-Top-1:} 
as a formal language corpus of texts on economics, we determine the subset of all Wikipedia articles whose top-level topic category corresponds to the \textit{Dewey Decimal Classification} (DDC) \textit{Category 330} (\textit{Economics}). 
In other words, we DDC-categorize all Wikipedia articles of the German Wikipedia using text2ddc \citep{Uslu:Mehler:Baumartz:2019} and select those articles whose top-level topic category corresponds to DDC category 330. 
In this way, we obtain a subset of Wikipedia articles that can be very reliably assigned to our target topic of economics: 
anyone who reads articles of the Wikipedia article network, which is spanned by these articles, navigates, so to speak, in the thematically homogeneous area of economically relevant articles.

\item \textit{Corpus WP-Top-3:} in analogy to WP-Top-1, WP-Top-3 is the set of all German Wikipedia articles where the DDC category 330 is among the first three DDC categories assigned to this article by text2ddc with a membership value of at least 10\%.
Obviously WP-Top-3 contains larger parts of WP-Top-1 (10\% threshold) or even this corpus as a whole, but likely also articles whose relation to economics is less confirmed, even if they do not fall below the 10\% threshold. 

\item \textit{Corpus WP-Eco:} 
WP-Eco is the corpus of all articles in Wikipedia that are directly or indirectly assigned to the category \textit{Wirtschaft} from Wikipedia's category system. 
WP-Eco contains \numprint{653397} articles and thus about a third of all \numprint{1760875} articles of German Wikipedia; WP-Eco also contains articles that are (possibly) only (very) indirectly related to the topic of economics. 
Whoever reads articles from the corresponding article network navigates, so to speak, in the wider area of economics-related articles, while possibly changing the topic (starting from economics), but in a frame that still has to do with economics.

\item \textit{Corpus WP-All:} 
the largest corpus we look at includes the \numprint{1760875} articles from the German Wikipedia, most of which are not related to economics (see Table \ref{tab:corpora}).

\item \textit{Corpus ZEIT:} 
as a second corpus of informal language of press communication, we process the \numprint{184186} texts of the German newspaper \textit{Die Zeit} published in the years 1994-2014.

%

\item \textit{Corpus TB:} Last but nor least we analyze a corpus of formal language, that is, a corpus of 14 textbooks all about economics in the narrow sense.
\end{enumerate}

In total, we consider eight corpora, three of which are informal language corpora of press communication (SZ-Eco, SZ-All, ZEIT), three of which mainly comprise texts that are not related to economics (SZ-All, WP-All, ZEIT) and five of which are formal language corpora (WP-All, WP-Eco, WP-Top-1, WP-Top-3, TB). 
Moreover, one of the informal language corpora (SZ-Eco) and four of the formal language corpora (WP-Eco, WP-Top-1, WP-Top-3, TB) focus more or less on economics. 
For preprocessing all these corpora, we use TextImager \citep{Hemati:Uslu:Mehler:2016}. That is, the corpora are tokenized, part of speech-tagged and lemmatized. 
Furthermore, sentences are split and tokens are segmented to identify candidate compounds, their heads and modifiers.
Text classification regarding the second level of the DDC is performed by means of text2ddc \citep{Uslu:Mehler:Baumartz:2019}.
Embeddings are computed for all corpora separately using word2vec based on standard settings (i.e.\ word vector size $=$ 100, window size $=$ 5, with 5 training iterations) for skip-gram and cbow (see \citealt{Mehler:et:al:2020b} for a related procedure).
Finally, the embeddings are used to induce TCNs according to Section \ref{sec:A Two-part Procedure for Measuring the Use of Threshold Concepts}, which are then processed with GraphMiner, a network analysis software under development at TTLab (\href{www.texttechnologylab.org}{www.texttechnologylab.org}).

\subsection{Results}
\label{subsec:results}

\begin{figure}[t]
    \centering
	\includegraphics[width=0.7\linewidth]{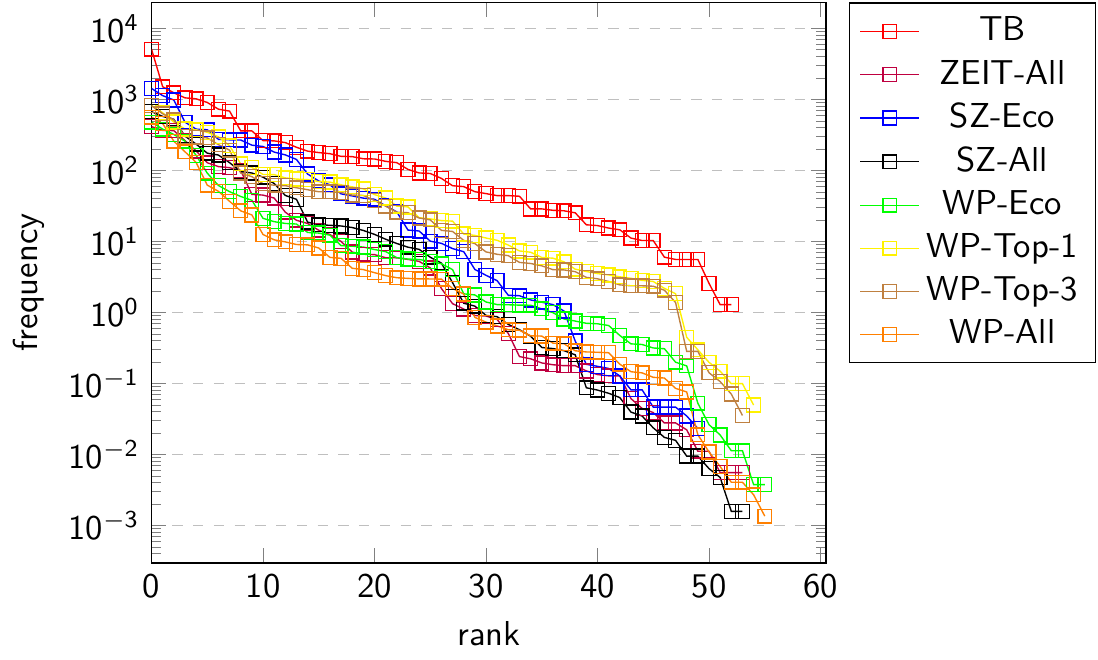}
\caption{Rank-specificity distribution of threshold concepts based on lemma (s-specificity) and compounding frequencies (c-specificity).}
\label{fig:RankSpecificityDistribution-All}
\end{figure}

In Figure \ref{fig:RankSpecificityDistribution-All} we show the rank-specificity distribution of our set of threshold concepts based on the variant $\mu$ of vertex weights in TCNs.
It is remarkable that the specificity values of threshold concepts in textbooks are above all distributions induced by the comparison corpora.
Furthermore, the specificity values for concepts from formal language corpora dedicated to economics such as WP-Top-1 and WP-Top-3 are also higher.
In contrast, specificity values from corpora of more general content (WP-All, SZ-All, ZEIT-All) do not achieve such high levels. 
In the middle of the spectrum of specificity distributions we observe SZ-Eco and WP-Eco, two corpora of medium size, which deal with economic issues in a larger thematic context.
Note that we calculate relative frequencies in order to rule out size effects and scale the distributions (by multiplying with \numprint{1000000}) in order to enhance readability. 

In order to estimate whether the distributions actually differ from each other, we perform pairwise Kolmogorov-Smirnov goodness-of-fit tests. If the $p$-values of any such fit is high, then we cannot reject the hypothesis that the distributions of the two samples are the same. In other words: small $p$-values indicate a significant difference between two distributions. Results are collected in Tab.~\ref{tab:ks-all}, where $p < 0.1$ is highlighted in green (likewise for Tabs.~\ref{tab:ks-only} and \ref{tab:ks-without} below):
obviously, in most cases the distributions differ from each other. Remarkable exceptions are SZ-Eco in relation to SZ-All (the latter contains the former), WP-Top-1 and WP-Top-3 (also a matter of inclusion) and especially SZ-Eco in relation to WP-All. 

\begin{table}[htb]
\nprounddigits{4}\footnotesize
\centering
\begin{adjustbox}{max width=\textwidth}
\begin{tabular}{lllllllll}
\toprule
{} &   TB &                      SZ-Eco &                      SZ-All &                      WP-Eco &                    WP-Top-1 &                    WP-Top-3 &                       WP-All &                  Zeit-All \\
\midrule
TB       &  --- &  \gn{0.0004708777137311104} &  \gn{7.242378898264512e-06} &  \gn{0.0017584914118611294} &    \gn{0.05780142152081069} &    \gn{0.04156131025853971} &  \gn{0.00033819378447963455} &  \gn{0.00671503163114473} \\
SZ-Eco   &  --- &                         --- &     \rn{0.4411357887480324} &     \rn{0.5834427748235623} &   \gn{0.015360311869018872} &    \gn{0.01807746883817274} &      \rn{0.8232981956014923} &   \rn{0.4411357887480324} \\
SZ-All   &  --- &                         --- &                         --- &    \gn{0.05310253294052336} &  \gn{0.0016666788642411001} &  \gn{0.0023073562280544638} &     \rn{0.21918754631932846} &  \gn{0.08735018691301626} \\
WP-Eco   &  --- &                         --- &                         --- &                         --- &    \rn{0.13016105854605098} &    \rn{0.15659904291764148} &      \rn{0.7789498597918985} &  \gn{0.03375833130600725} \\
WP-Top-1 &  --- &                         --- &                         --- &                         --- &                         --- &     \rn{0.9615099726850278} &    \gn{0.008568230089663231} &  \gn{0.06872200206455337} \\
WP-Top-3 &  --- &                         --- &                         --- &                         --- &                         --- &                         --- &    \gn{0.011235687099652725} &  \gn{0.05266941344461325} \\
WP-All   &  --- &                         --- &                         --- &                         --- &                         --- &                         --- &                          --- &  \gn{0.03375833130600725} \\
Zeit-All &  --- &                         --- &                         --- &                         --- &                         --- &                         --- &                          --- &                       --- \\
\bottomrule
\end{tabular}
    
\end{adjustbox}
\caption{$P$-values of the Kolmogorov-Smirnov goodness-of-fit test applied to the pairwise combinations of the distributions in Fig.~\protect\ref{fig:RankSpecificityDistribution-All}.}
    \label{tab:ks-all}
\end{table}

\begin{table}[htb]
\nprounddigits{4}\footnotesize
\centering
\begin{adjustbox}{max width=\textwidth}
\begin{tabular}{lllllllll}
\toprule
{} &   TB &                       SZ-Eco &                       SZ-All &                      WP-Eco &                    WP-Top-1 &                    WP-Top-3 &                       WP-All &                   Zeit-All \\
\midrule
TB       &  --- &  \gn{0.00034931906804602786} &  \gn{1.0799550135720537e-05} &  \gn{0.0005032588922944115} &     \rn{0.1384130999542914} &    \gn{0.04342248866188758} &  \gn{0.00010001139203774656} &  \gn{0.007085381298799098} \\
SZ-Eco   &  --- &                          --- &      \rn{0.5626692639126882} &     \rn{0.1840454723690219} &   \gn{0.013233141313409358} &   \gn{0.013519326658616548} &      \rn{0.5388188754133296} &    \rn{0.2330140089842796} \\
SZ-All   &  --- &                          --- &                          --- &   \gn{0.014960939814501861} &  \gn{0.0002199512587414132} &  \gn{0.0014651981669114855} &     \gn{0.07502552440059829} &  \gn{0.023159628280428723} \\
WP-Eco   &  --- &                          --- &                          --- &                         --- &    \gn{0.04908325004247327} &    \gn{0.09647617531948227} &      \rn{0.4681496145911523} &   \gn{0.09951968254656018} \\
WP-Top-1 &  --- &                          --- &                          --- &                         --- &                         --- &     \rn{0.9665827497829119} &    \gn{0.008076547801333933} &   \gn{0.06949117791594761} \\
WP-Top-3 &  --- &                          --- &                          --- &                         --- &                         --- &                         --- &    \gn{0.005648344917303372} &   \gn{0.04750001877647403} \\
WP-All   &  --- &                          --- &                          --- &                         --- &                         --- &                         --- &                          --- &   \gn{0.06188429212686586} \\
Zeit-All &  --- &                          --- &                          --- &                         --- &                         --- &                         --- &                          --- &                        --- \\
\bottomrule
\end{tabular}
    
\end{adjustbox}
\caption{$P$-values of the Kolmogorov-Smirnov goodness-of-fit test applied to the pairwise combinations of the distributions in Fig.~\protect\ref{fig:RankSpecificityDistribution-LemmaFreqOnly} (s-specificity).}
    \label{tab:ks-only}
\end{table}

\begin{table}[htb]
\nprounddigits{4}\footnotesize
\centering
\begin{adjustbox}{max width=\textwidth}
\begin{tabular}{lllllllll}
\toprule
{} &   TB &                     SZ-Eco &                     SZ-All &                    WP-Eco &                  WP-Top-1 &                  WP-Top-3 &                     WP-All &                  Zeit-All \\
\midrule
TB       &  --- &  \gn{0.010806760526342885} &  \gn{0.000687713620405983} &  \gn{0.03929830237412535} &  \rn{0.21840720056444463} &  \rn{0.18033703360653552} &  \gn{0.012234258826202438} &  \gn{0.06644979307220777} \\
SZ-Eco   &  --- &                        --- &    \rn{0.3136253290192914} &   \rn{0.6937356011009224} &   \rn{0.1941234474668937} &   \rn{0.2435852237010354} &    \rn{0.9955815254572036} &   \rn{0.5978448980687857} \\
SZ-All   &  --- &                        --- &                        --- &  \rn{0.20152189449376567} &  \gn{0.02253816727705038} &  \gn{0.05266941344461325} &     \rn{0.650513214900474} &  \rn{0.13931665637899754} \\
WP-Eco   &  --- &                        --- &                        --- &                       --- &  \rn{0.43541647199783917} &    \rn{0.592951966386642} &    \rn{0.9807333238855117} &   \rn{0.4842652416535621} \\
WP-Top-1 &  --- &                        --- &                        --- &                       --- &                       --- &   \rn{0.9961994591104391} &   \rn{0.18805565064284946} &   \rn{0.5259270525856905} \\
WP-Top-3 &  --- &                        --- &                        --- &                       --- &                       --- &                       --- &    \rn{0.3078726612319459} &   \rn{0.5984842563735271} \\
WP-All   &  --- &                        --- &                        --- &                       --- &                       --- &                       --- &                        --- &  \rn{0.29854027039532194} \\
Zeit-All &  --- &                        --- &                        --- &                       --- &                       --- &                       --- &                        --- &                       --- \\
\bottomrule
\end{tabular}
    
\end{adjustbox}
\caption{$P$-values of the Kolmogorov-Smirnov goodness-of-fit test applied to the pairwise combinations of the distributions in Fig.~\protect\ref{fig:RankSpecificityDistribution-CompoundingFreqOnly}  (c-specificity).}
    \label{tab:ks-without}
\end{table}

\begin{figure}[t]
    \centering
	\includegraphics[width=0.7\linewidth]{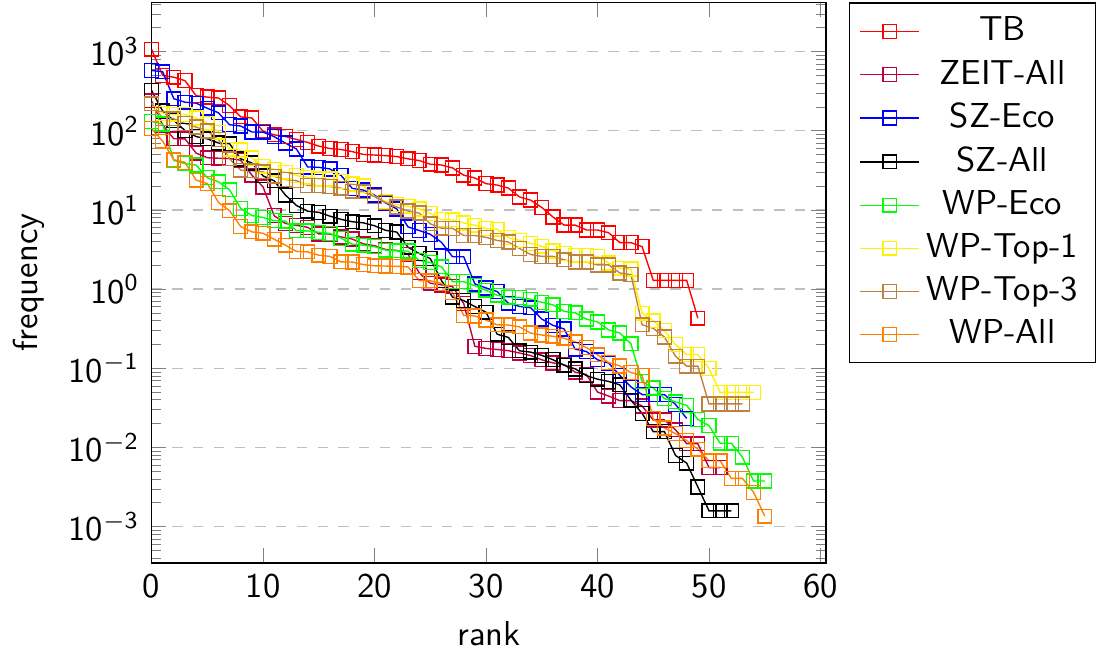}
\caption{Rank-specificity distribution of threshold concepts based on lemma frequencies.}
\label{fig:RankSpecificityDistribution-LemmaFreqOnly}
\end{figure}

\begin{figure}[t]
    \centering
	\includegraphics[width=0.7\linewidth]{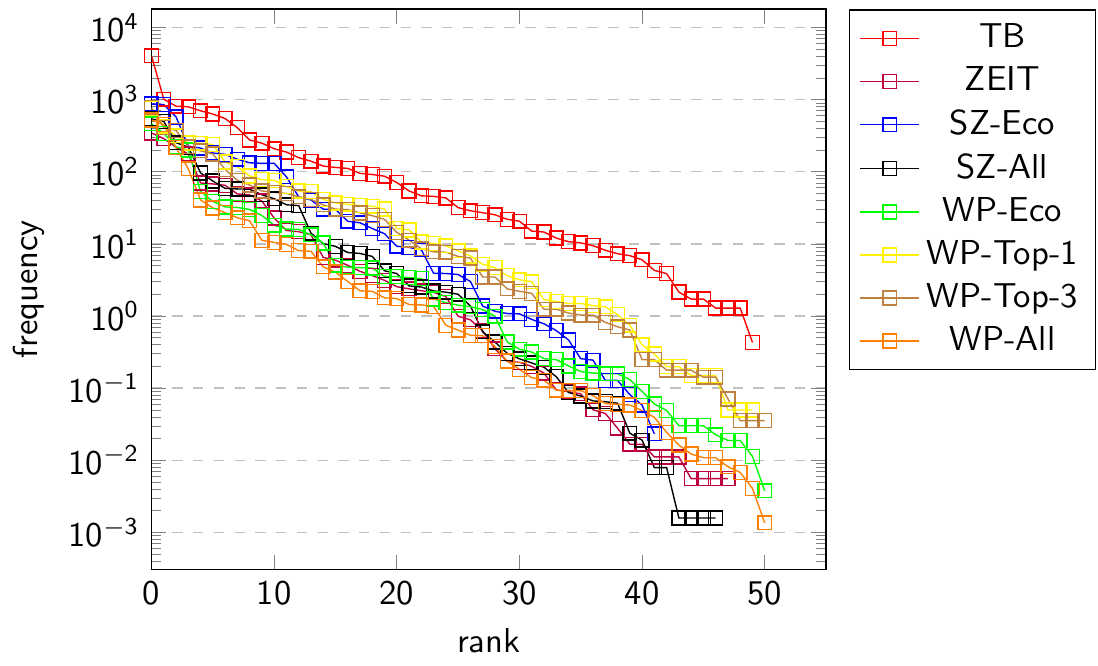}
\caption{Rank-specificity distribution of threshold concepts based on compounding frequencies.}
\label{fig:RankSpecificityDistribution-CompoundingFreqOnly}
\end{figure}

The scenario observed in Figure \ref{fig:RankSpecificityDistribution-All} is also displayed by Figure  
\ref{fig:RankSpecificityDistribution-LemmaFreqOnly} (s-specificity) and Figure \ref{fig:RankSpecificityDistribution-CompoundingFreqOnly} (c-specificity): 
the specificity distributions are all topped by the distribution for textbooks.
In this sense, it can be said that the threshold concepts considered here are most specifically described in the formal language textbook corpus, followed by the two formal language Wikipedia-based corpora WP-Top-1 and WP-Top-3 and least specifically in the informal newspaper corpora SZ-All and ZEIT-All, although in the case of c-specificity the situation is not so obvious.
A borderline case is WP-Eco, a corpus that consists of Wikipedia articles that are directly or indirectly assigned to the thematic field of economics.
%

When we look at tables \ref{tab:ks-only} and \ref{tab:ks-without}, we get the information that while the frequency distributions (s-specificity) tend to be distinguishable, the distinguishability of the c-specificities is much less: obviously, the frequencies of compounds to which our threshold concepts belong are more independent of the underlying corpus.
Moreover, the distributions in Figure \ref{fig:RankSpecificityDistribution-All}--\ref{fig:RankSpecificityDistribution-CompoundingFreqOnly} tend to be all Zipfian:
although a lognormal distribution is also a good fit in 17 (of 24) cases, power law fitting is still a valid option (there is not a single significant $p$-value $<0.05$ for any $R < 0$; note further that a lognormal distribution is a heavy-tailed distribution, too):
the exponent $\alpha$ ranges from $\approx 1.3$ to $\approx 2.8$, where the minimum $x$ value of the fit is given as \enquote{x-min}, 
see Table \ref{tab:Power-law-fitting}.\footnote{We apply the toolbox of \citet{Alstott:Bullmore:Plenz:2014} according to \citet{Clauset:Shalizi:Newman:2009}: 
power laws (first) are compared to lognormal distributions (second): 
\enquote{$R$ is the loglikelihood ratio between the two candidate distributions. This number will be positive if the data is more likely in the first distribution, and negative if the data is more likely in the second distribution. 
The significance value for that direction is $p$.} \citep[5]{Alstott:Bullmore:Plenz:2014}}
\begin{table}[htb]
    \centering\footnotesize
    \textbf{lemma and compound}, Fig.~\protect\ref{fig:RankSpecificityDistribution-All} \par
    \begin{tabular}{lrrrr}
\toprule
{} &     alpha &  x-min &         R &         P \\
\midrule
SZ-All   &  1.711629 &    4.0 &  0.004996 &  0.737173 \\
SZ-Eco   &  1.565592 &    2.0 & -0.686457 &  0.305550 \\
TB       &  2.825375 &   41.0 &  0.009203 &  0.853003 \\
WP-All   &  1.593086 &    3.0 &  0.009178 &  0.719042 \\
WP-Eco   &  1.613395 &    3.0 & -0.568429 &  0.322225 \\
WP-Top-1 &  1.522315 &    3.0 & -0.002717 &  0.793323 \\
WP-Top-3 &  1.485334 &    2.0 & -0.242887 &  0.866997 \\
Zeit-All &  1.395048 &    1.0 & -0.666070 &  0.403735 \\
\bottomrule
\end{tabular}

    \par\bigskip
    \textbf{only lemma}, Fig.~\protect\ref{fig:RankSpecificityDistribution-LemmaFreqOnly} \par
    \begin{tabular}{lrrrr}
\toprule
{} &     alpha &  x-min &         R &         P \\
\midrule
SZ-All   &  1.735390 &    5.0 & -0.126319 &  0.294422 \\
SZ-Eco   &  1.548944 &    2.0 & -0.689713 &  0.307030 \\
TB       &  1.336778 &    1.0 & -0.627145 &  0.792154 \\
WP-All   &  1.593086 &    3.0 &  0.009178 &  0.719042 \\
WP-Eco   &  1.483664 &    2.0 &  0.000933 &  0.964148 \\
WP-Top-1 &  1.433428 &    2.0 & -0.173451 &  0.273327 \\
WP-Top-3 &  1.546151 &    4.0 &  0.015474 &  0.301308 \\
Zeit-All &  1.395048 &    1.0 & -0.666070 &  0.403735 \\
\bottomrule
\end{tabular}

    \par\bigskip
    \textbf{only compound}, Fig.~\protect\ref{fig:RankSpecificityDistribution-CompoundingFreqOnly}\par
    \begin{tabular}{lrrrr}
\toprule
{} &     alpha &  x-min &         R &         P \\
\midrule
SZ-All   &  1.355914 &    1.0 & -0.666272 &  0.268328 \\
SZ-Eco   &  1.559961 &    3.0 &  0.028074 &  0.235693 \\
TB       &  1.395212 &    2.0 & -0.683652 &  0.307499 \\
WP-All   &  1.395048 &    1.0 & -0.666070 &  0.403735 \\
WP-Eco   &  1.676775 &    4.0 & -0.580707 &  0.369533 \\
WP-Top-1 &  1.545465 &    3.0 & -0.641716 &  0.320173 \\
WP-Top-3 &  1.464465 &    2.0 & -0.649692 &  0.317607 \\
Zeit-All &  1.418524 &    2.0 & -0.080603 &  0.606211 \\
\bottomrule
\end{tabular}

    \caption{Power law goodness-of-fit tests for the distributions from Figs.~\protect\ref{fig:RankSpecificityDistribution-All}, \protect\ref{fig:RankSpecificityDistribution-LemmaFreqOnly}, and \protect\ref{fig:RankSpecificityDistribution-CompoundingFreqOnly}.}
    \label{tab:Power-law-fitting}
\end{table}
%
%
From this perspective, we see the alternative of hypothesis H1a, which states that the use of threshold concepts in formal language corpora is neither more c-specific nor more s-specific than in informal language corpora, as being falsified.

\begin{table}[htb]
\nprounddigits{4}\footnotesize
\centering
\textbf{$x$-values}:\par 
\begin{adjustbox}{max width=\textwidth}
\begin{tabular}{lllllllll}
\toprule
{} & SZ-All &                       SZ-Eco &                           TB &                       WP-All &                       WP-Eco &                     WP-Top-1 &                      WP-Top-3 &                     Zeit-All \\
\midrule
SZ-All   &    --- &  \gn{1.3322676295501878e-15} &   \gn{6.661338147750939e-16} &   \gn{3.774758283725532e-15} &   \gn{3.774758283725532e-15} &  \gn{3.254679608977452e-300} &  \gn{2.1646377602758511e-299} &  \gn{1.4432899320127035e-15} \\
SZ-Eco   &    --- &                          --- &  \gn{2.1329858606811136e-05} &  \gn{1.5543122344752192e-15} &  \gn{1.5543122344752192e-15} &  \gn{1.3322676295501878e-15} &   \gn{1.3322676295501878e-15} &  \gn{1.1102230246251565e-16} \\
TB       &    --- &                          --- &                          --- &  \gn{1.5543122344752192e-15} &  \gn{1.5543122344752192e-15} &   \gn{6.661338147750939e-16} &    \gn{6.661338147750939e-16} &  \gn{1.5543122344752192e-15} \\
WP-All   &    --- &                          --- &                          --- &                          --- &      \rn{0.3046738960267182} &   \gn{9.832576356449962e-07} &    \gn{0.0002125731914102147} &  \gn{2.1094237467877974e-15} \\
WP-Eco   &    --- &                          --- &                          --- &                          --- &                          --- &  \gn{1.7187396805784516e-07} &    \gn{6.944014693321954e-05} &  \gn{2.1094237467877974e-15} \\
WP-Top-1 &    --- &                          --- &                          --- &                          --- &                          --- &                          --- &      \rn{0.47749460966697926} &  \gn{1.4432899320127035e-15} \\
WP-Top-3 &    --- &                          --- &                          --- &                          --- &                          --- &                          --- &                           --- &  \gn{1.4432899320127035e-15} \\
Zeit-All &    --- &                          --- &                          --- &                          --- &                          --- &                          --- &                           --- &                          --- \\
\bottomrule
\end{tabular}
    
\end{adjustbox}
\par\bigskip
\textbf{$y$-values}:\par 
\nprounddigits{2}
\begin{adjustbox}{max width=\textwidth}
\begin{tabular}{lllllllll}
\toprule
{} & SZ-All &    SZ-Eco &        TB &    WP-All &    WP-Eco &  WP-Top-1 &  WP-Top-3 &  Zeit-All \\
\midrule
SZ-All   &    --- &  \rn{1.0} &  \rn{1.0} &  \rn{1.0} &  \rn{1.0} &  \rn{1.0} &  \rn{1.0} &  \rn{1.0} \\
SZ-Eco   &    --- &       --- &  \rn{1.0} &  \rn{1.0} &  \rn{1.0} &  \rn{1.0} &  \rn{1.0} &  \rn{1.0} \\
TB       &    --- &       --- &       --- &  \rn{1.0} &  \rn{1.0} &  \rn{1.0} &  \rn{1.0} &  \rn{1.0} \\
WP-All   &    --- &       --- &       --- &       --- &  \rn{1.0} &  \rn{1.0} &  \rn{1.0} &  \rn{1.0} \\
WP-Eco   &    --- &       --- &       --- &       --- &       --- &  \rn{1.0} &  \rn{1.0} &  \rn{1.0} \\
WP-Top-1 &    --- &       --- &       --- &       --- &       --- &       --- &  \rn{1.0} &  \rn{1.0} \\
WP-Top-3 &    --- &       --- &       --- &       --- &       --- &       --- &       --- &  \rn{1.0} \\
Zeit-All &    --- &       --- &       --- &       --- &       --- &       --- &       --- &       --- \\
\bottomrule
\end{tabular}
    
\end{adjustbox}
\caption{$P$-values of the Kolmogorov-Smirnov goodness-of-fit test applied to the pairwise combinations of the $x$ and $y$ values of the distributions in Fig.~\protect\ref{fig:MinEdgeWeightVsCohesion}.}
    \label{tab:cohesion-x}
\end{table}


\begin{table}[htb]
\nprounddigits{4}\footnotesize
\centering
\textbf{$x$-values}:\par 
\begin{adjustbox}{max width=\textwidth}
\begin{tabular}{lllllllll}
\toprule
{} & SZ-All &                       SZ-Eco &                           TB &                       WP-All &                       WP-Eco &                     WP-Top-1 &                      WP-Top-3 &                     Zeit-All \\
\midrule
SZ-All   &    --- &  \gn{1.3322676295501878e-15} &   \gn{6.661338147750939e-16} &   \gn{3.774758283725532e-15} &   \gn{3.774758283725532e-15} &  \gn{3.254679608977452e-300} &  \gn{2.1646377602758511e-299} &  \gn{1.4432899320127035e-15} \\
SZ-Eco   &    --- &                          --- &  \gn{2.1329858606811136e-05} &  \gn{1.5543122344752192e-15} &  \gn{1.5543122344752192e-15} &  \gn{1.3322676295501878e-15} &   \gn{1.3322676295501878e-15} &  \gn{1.1102230246251565e-16} \\
TB       &    --- &                          --- &                          --- &  \gn{1.5543122344752192e-15} &  \gn{1.5543122344752192e-15} &   \gn{6.661338147750939e-16} &    \gn{6.661338147750939e-16} &  \gn{1.5543122344752192e-15} \\
WP-All   &    --- &                          --- &                          --- &                          --- &      \rn{0.3046738960267182} &   \gn{9.832576356449962e-07} &    \gn{0.0002125731914102147} &  \gn{2.1094237467877974e-15} \\
WP-Eco   &    --- &                          --- &                          --- &                          --- &                          --- &  \gn{1.7187396805784516e-07} &    \gn{6.944014693321954e-05} &  \gn{2.1094237467877974e-15} \\
WP-Top-1 &    --- &                          --- &                          --- &                          --- &                          --- &                          --- &      \rn{0.47749460966697926} &  \gn{1.4432899320127035e-15} \\
WP-Top-3 &    --- &                          --- &                          --- &                          --- &                          --- &                          --- &                           --- &  \gn{1.4432899320127035e-15} \\
Zeit-All &    --- &                          --- &                          --- &                          --- &                          --- &                          --- &                           --- &                          --- \\
\bottomrule
\end{tabular}
    
\end{adjustbox}
\par\bigskip
\textbf{$y$-values}:\par
\begin{adjustbox}{max width=\textwidth}
\begin{tabular}{lllllllll}
\toprule
{} & SZ-All &    SZ-Eco &                           TB &                       WP-All &                       WP-Eco &                     WP-Top-1 &                     WP-Top-3 &                     Zeit-All \\
\midrule
SZ-All   &    --- &  \rn{1.0} &   \gn{6.661338147750939e-16} &   \gn{3.774758283725532e-15} &   \gn{3.774758283725532e-15} &  \gn{1.0863945026404866e-54} &    \gn{2.44527920265294e-71} &  \gn{1.4432899320127035e-15} \\
SZ-Eco   &    --- &       --- &  \gn{1.2212453270876722e-15} &  \gn{1.5543122344752192e-15} &  \gn{1.5543122344752192e-15} &  \gn{1.3322676295501878e-15} &  \gn{1.3322676295501878e-15} &  \gn{1.1102230246251565e-16} \\
TB       &    --- &       --- &                          --- &  \gn{1.1767359309189374e-09} &   \gn{5.162537064506978e-14} &   \gn{1.477740152466822e-11} &   \gn{4.872130476840653e-10} &   \gn{1.767256341267398e-10} \\
WP-All   &    --- &       --- &                          --- &                          --- &   \gn{6.728102756458281e-05} &  \gn{3.7346411701721927e-05} &    \gn{0.012577072026717317} &   \gn{7.908784738219765e-12} \\
WP-Eco   &    --- &       --- &                          --- &                          --- &                          --- &  \gn{4.0788483701703626e-11} &   \gn{3.815155427133732e-06} &  \gn{2.1094237467877974e-15} \\
WP-Top-1 &    --- &       --- &                          --- &                          --- &                          --- &                          --- &   \gn{0.0009305757718496391} &   \gn{1.042055330913172e-12} \\
WP-Top-3 &    --- &       --- &                          --- &                          --- &                          --- &                          --- &                          --- &  \gn{3.1530333899354446e-14} \\
Zeit-All &    --- &       --- &                          --- &                          --- &                          --- &                          --- &                          --- &                          --- \\
\bottomrule
\end{tabular}
    
\end{adjustbox}
\caption{$P$-values of the Kolmogorov-Smirnov goodness-of-fit test applied to the pairwise combinations of the $x$ and $y$ values of the distributions in Fig.~\protect\ref{fig:MinEdgeWeightVsCWS}.}
    \label{tab:cws-x}
\end{table}


\begin{table}[htb]
\nprounddigits{4}\footnotesize
\centering
\textbf{$x$-values}:\par
\begin{adjustbox}{max width=\textwidth}
\begin{tabular}{lllllllll}
\toprule
{} & SZ-All &                       SZ-Eco &                           TB &                       WP-All &                       WP-Eco &                     WP-Top-1 &                     WP-Top-3 &                     Zeit-All \\
\midrule
SZ-All   &    --- &  \gn{1.3322676295501878e-15} &  \gn{1.7763568394002505e-15} &   \gn{3.774758283725532e-15} &   \gn{3.774758283725532e-15} &   \gn{3.835582941962095e-41} &  \gn{2.1932185140276563e-33} &     \rn{0.19984026608811756} \\
SZ-Eco   &    --- &                          --- &  \gn{5.5968341072798466e-11} &  \gn{1.5543122344752192e-15} &  \gn{1.5543122344752192e-15} &  \gn{1.3322676295501878e-15} &  \gn{1.3322676295501878e-15} &  \gn{1.1102230246251565e-16} \\
TB       &    --- &                          --- &                          --- &  \gn{1.5543122344752192e-15} &  \gn{1.5543122344752192e-15} &   \gn{6.661338147750939e-16} &   \gn{6.661338147750939e-16} &  \gn{3.1086244689504383e-15} \\
WP-All   &    --- &                          --- &                          --- &                          --- &    \gn{0.000564531340349513} &      \rn{0.5953083448282726} &     \gn{0.04185817856648544} &  \gn{2.1094237467877974e-15} \\
WP-Eco   &    --- &                          --- &                          --- &                          --- &                          --- &   \gn{0.0016729729960333062} &     \rn{0.47164894877017227} &  \gn{2.1094237467877974e-15} \\
WP-Top-1 &    --- &                          --- &                          --- &                          --- &                          --- &                          --- &     \gn{0.04722432602160368} &  \gn{1.4432899320127035e-15} \\
WP-Top-3 &    --- &                          --- &                          --- &                          --- &                          --- &                          --- &                          --- &  \gn{1.4432899320127035e-15} \\
Zeit-All &    --- &                          --- &                          --- &                          --- &                          --- &                          --- &                          --- &                          --- \\
\bottomrule
\end{tabular}
    
\end{adjustbox}
\par\bigskip
\textbf{$y$-values}:\par
\nprounddigits{2}
\begin{adjustbox}{max width=\textwidth}
\begin{tabular}{lllllllll}
\toprule
{} & SZ-All &    SZ-Eco &        TB &    WP-All &    WP-Eco &  WP-Top-1 &  WP-Top-3 &  Zeit-All \\
\midrule
SZ-All   &    --- &  \rn{1.0} &  \rn{1.0} &  \rn{1.0} &  \rn{1.0} &  \rn{1.0} &  \rn{1.0} &  \rn{1.0} \\
SZ-Eco   &    --- &       --- &  \rn{1.0} &  \rn{1.0} &  \rn{1.0} &  \rn{1.0} &  \rn{1.0} &  \rn{1.0} \\
TB       &    --- &       --- &       --- &  \rn{1.0} &  \rn{1.0} &  \rn{1.0} &  \rn{1.0} &  \rn{1.0} \\
WP-All   &    --- &       --- &       --- &       --- &  \rn{1.0} &  \rn{1.0} &  \rn{1.0} &  \rn{1.0} \\
WP-Eco   &    --- &       --- &       --- &       --- &       --- &  \rn{1.0} &  \rn{1.0} &  \rn{1.0} \\
WP-Top-1 &    --- &       --- &       --- &       --- &       --- &       --- &  \rn{1.0} &  \rn{1.0} \\
WP-Top-3 &    --- &       --- &       --- &       --- &       --- &       --- &       --- &  \rn{1.0} \\
Zeit-All &    --- &       --- &       --- &       --- &       --- &       --- &       --- &       --- \\
\bottomrule
\end{tabular}
    
\end{adjustbox}
\caption{$P$-values of the Kolmogorov-Smirnov goodness-of-fit test applied to the pairwise combinations of the $x$ and $y$ values of the distributions in Fig.~\protect\ref{fig:MinEdgeWeightVsCohesionCBOW}.}
    \label{tab:cohesion-cbow-y}
\end{table}

\begin{figure}[t]
    \centering
	\includegraphics[width=0.7\linewidth]{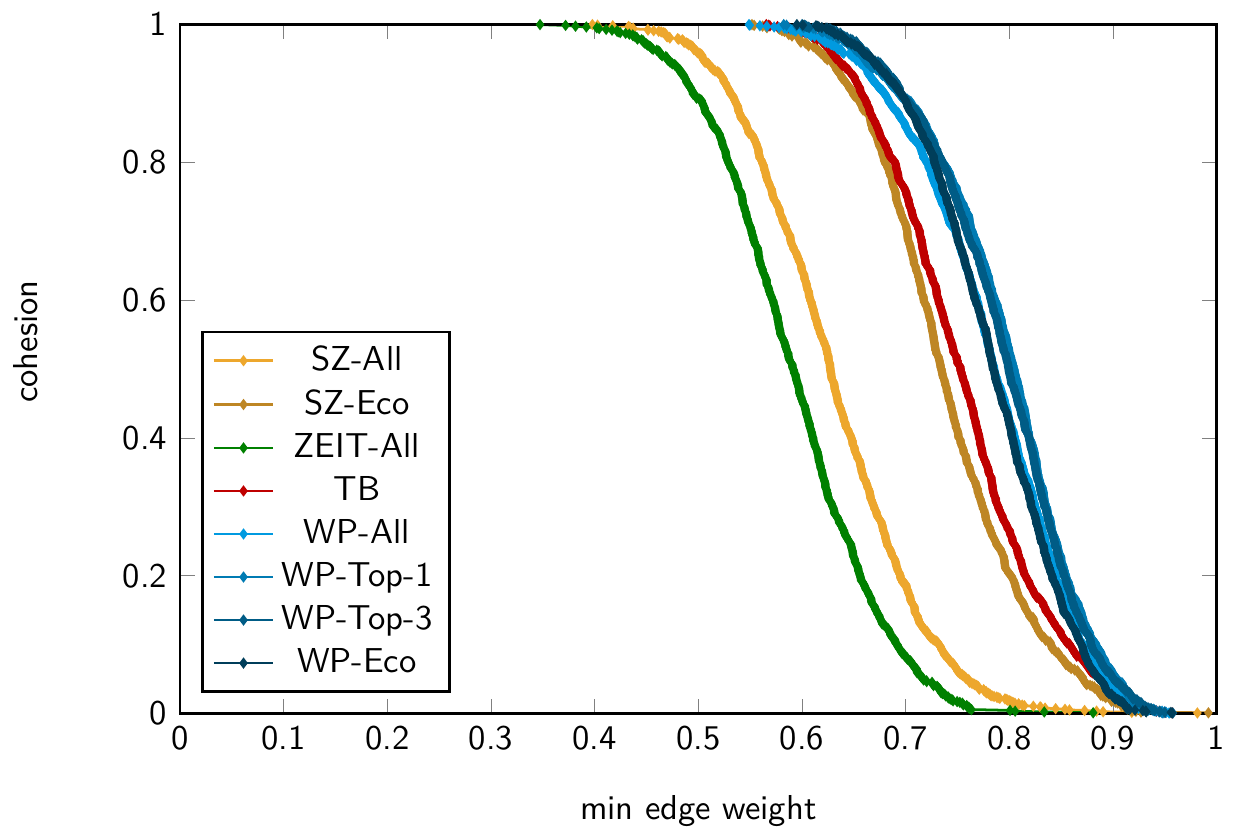}
\caption{Network cohesion as a function of the minimum weight per $\alpha$-cut of word embedding networks of threshold concepts according to the skip-gram model.}
\label{fig:MinEdgeWeightVsCohesion}
\end{figure}

Next we consider Hypothesis H1b.
For this purpose, we compare the series of cohesion values induced by the series of alpha-cut graphs (see above) based on our eight different corpora: 
Figure \ref{fig:MinEdgeWeightVsCohesion} shows the corresponding distributions starting from the TCNs derived from word embedding similarities based on the skip-gram model of word2vec and thus for syntagmatic associations (starting from the respective seed word to the probable context in the sense of being defined by neighboring words).
Very remarkably, all four Wikipedia corpora behave very alike:
the cohesion values of the TCN series induced by these corpora decrease at the latest compared to all other corpora and their corresponding TCN series, that is, they decrease for the comparatively highest $\alpha$ values.
Conversely, the cohesion values of the corresponding TCN series induced by the newspaper corpora (SZ-All, ZEIT-All) decrease the fastest.
In the middle of this spectrum we surprisingly observe two series of cohesion values: 
that for the textbook corpus and that for the economics-related SZ-Eco corpus, though rather in the neighborhood of the Wikipedia corpora than in the one of the newspaper corpora.
At this point, we have to ask whether the distributions shown in Figure \ref{fig:MinEdgeWeightVsCohesion} are actually different or not.
For this purpose we again perform Kolmogorov-Smirnov tests of goodness-of-fit, but now separately for both axes from Figure \ref{fig:MinEdgeWeightVsCohesion}. The reason is that neither axis is ordinally scaled, so we first perform a corresponding scaling before we can compare the corresponding feature distributions.
As shown in Table \ref{tab:cohesion-x}, we get a mixed result: while the alpha-cuts of the individual distributions increase very differently (so that the distributions are mostly clearly distinguishable from each other), this does not apply to the decreases in cohesion values caused by the increasing alpha-cuts: here the distributions are all indistinguishable.
For the distributions of the cohesion values this means that they are in fact all almost "identical" and therefore indistinguishable mirrored S-curves when being scaled appropriately.

From this spectrum of distributions, we get the following assessment: 
in Wikipedia-based corpora, the threshold concepts are most strongly associated with each other -- metaphorically speaking, they form a denser network of particles that are located much closer to each other. 
For much higher values than for any other corpus, the network cohesion (starting from a completely connected graph) takes a maximum value of 1;
and for equally maximum values the cohesion is at least 50\%, 75\% etc.:
the deletion of lower weighted edges in TCNs based on Wikipedia corpora is therefore more likely to lead to more cohesive networks compared to the other TCNs.
In view of this finding, the textbook-based TCNs are surprisingly less cohesive.
\textit{Based on our cognitive model, this suggests that reading such textbooks makes stronger syntagmatic associations under threshold concepts less likely.} 
Wikipedia seems to write more densely about these concepts, in a way that makes their associations more probable and also more pronounced. 
This may be related to the text type of Wikipedia (\textit{encyclopedic communication}) as opposed to textbooks, which may also contain longer motivational, exemplary or elaborating text passages. 
In any case, however, we see the hypothesis confirmed that formal language corpora make stronger associations between threshold concepts more likely than informal language corpora -- this is indirectly confirmed by the values of Table \ref{tab:cohesion-x} regarding the x-axis (formal language corpora are significantly \enquote{shifted} to the right compared to their newspaper-based counterparts, i.e.\ SZ-All and ZEIT-All).
An extreme-value-forming special position of textbooks, however, cannot be confirmed. 
Moreover, the strengths of the associations of threshold concepts obtained by means of informal texts on topics related to economics (SZ-Eco) can hardly be distinguished from those obtained with the help of textbooks:
\textit{from this point of view, we do not see a special role for textbooks compared to quasi informal newspaper articles}. 
The only exception is Wikipedia -- regardless of the topic of economics.

\begin{figure}[t]
    \centering
	\includegraphics[width=0.7\linewidth]{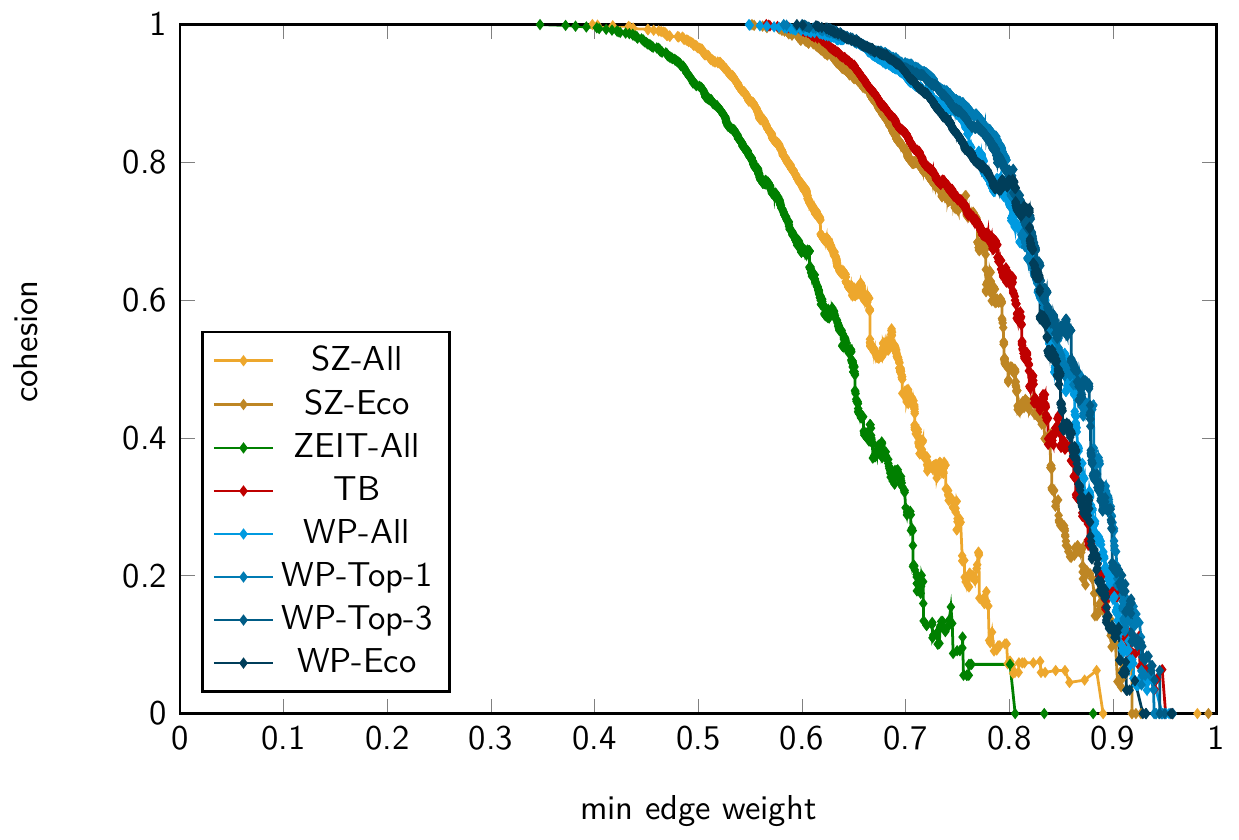} 
\caption{Network clustering per $\alpha$-cut ($\alpha$ = minimal allowable edge weight) of word embedding networks of threshold concepts (TCNs) based on the skip-gram model.}
\label{fig:MinEdgeWeightVsCWS}
\end{figure}

Figure \ref{fig:MinEdgeWeightVsCWS} essentially confirms the results obtained so far.
However, we now observe, for higher $\alpha$ values, that the cluster values of textbook-based networks become seemingly indistinguishable from those observable for Wikipedia corpora-based networks -- the same observation concerns the SZ-Eco-based networks.
Textbook-based TCNs are again hardly distinguishable from TCNs derived from informal language newspaper articles about topics related to economics (SZ-Eco). 
In any case, Table \ref{tab:cws-x} also shows that all value distributions along the x and y axis are now distinguishable with only three exceptions: the dynamics of clustering is obviously more corpus specific.

Any special role of textbooks almost completely disappears if we consider the cbow model of word2vec (i.e. associations starting from lexical contexts towards target words and thus paradigmatic associations) (see Figure \ref{fig:MinEdgeWeightVsCohesion}).
In other words, paradigmatic associations of the sort \textit{Bruttoinlandsprodukt}/\textit{gross domestic product} and \textit{BIP}/\textit{GDP} seem to be highest from the perspective of Wikipedia-based corpora and higher from the perspective of newspaper corpora than from the perspective of the textbook corpus, while syntagmatic associations of the sort \textit{Gewinn}/\textit{profit} and \textit{marginal}/\textit{marginal} are still highest in the case of Wikipedia-based corpora, but are more pronounced from the perspective of textbooks than from newspapers.
Table \ref{tab:cohesion-cbow-y} leads to an assessment similar to Table \ref{tab:cohesion-x}.

Note that in all these cases of cbow and skip-gram-based networks and their underlying embeddings we use standard parameter settings and especially a rate of 5 iterations:
from this point of view, it could be that shorter corpora are more negatively affected by such iterations than longer ones. 
Scaling their size by increasing the number of iterations can lead to false dissociations of words (as a test of 100 iterations based on the textbook corpus actually suggests). 
Instead, the sizes of the larger corpora should be reduced to those of the smallest corpora, that is, the corpus of textbooks -- but the corresponding sampling routine and experimentation will be part of future work. 
In any case, it should be noted that our results are conditioned by the latter assessment. 
And this means that the alternative of Hypothesis H1b is only falsified if we compare Wikipedia-based corpora with newspaper corpora. 
However, in the case of WP-All, we must refrain from a focus on economics-related topics.
The inclusion of the textbook corpus in the set of formal language corpora definitely does not allow such a falsification: 
so either H1b is wrong or our current measuring procedure does not allow yet for falsifying the alternative of H1b.

\begin{figure}[t]
    \centering
	\includegraphics[width=0.7\linewidth]{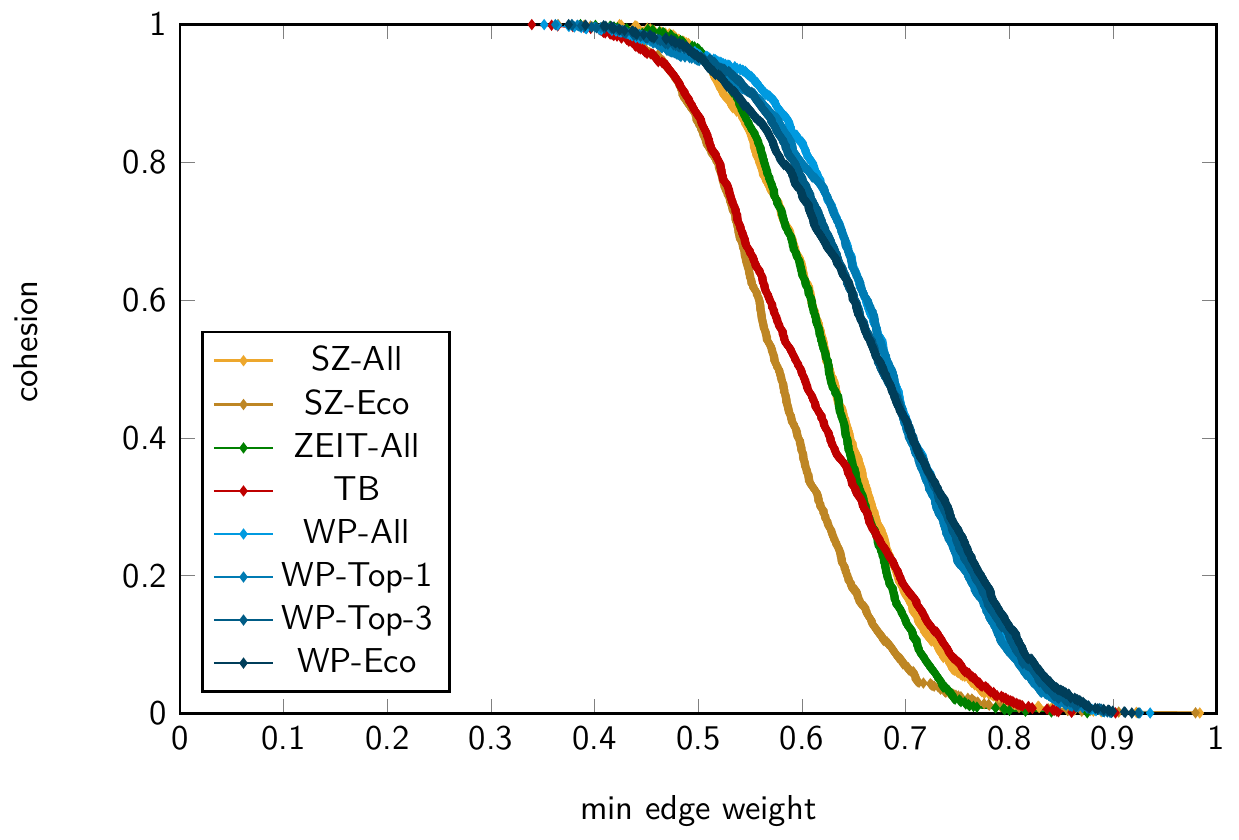} 
\caption{Network clustering per $\alpha$-cut ($\alpha$ = minimal allowable edge weight) of word embedding networks of threshold concepts (TCNs) based on the cbow model.}
\label{fig:MinEdgeWeightVsCohesionCBOW}
\end{figure}

\subsection{Discussion}
\label{subsec:discussion}

Section \ref{subsec:results} has shown that threshold concepts occur significantly more frequently in formal textbook corpora, both with respect to the naming variants investigated here and with respect to their frequencies as components of compounds: 
according to Hypothesis H1a, textbook corpora are more c- and s-specific than all other corpora investigated here.
However, we have also shown that their networking (according to stronger syntagmatic and paradigmatic associations) is not exceptional. 
What we observe as exceptional in this respect is Wikipedia, and this observation is independent of the topic of economics. 
This finding points to a special role of encyclopedic communication as a representative of formal language communication, a role that may have been underestimated in educational sciences until now. 
However, based on our experiments we must also note that we could not confirm H1b (or falsify its alternative hypothesis).

There are several points of departure for improving the procedure we have developed for measuring the usage regularities of threshold concepts in corpora of formal and informal language:
\begin{enumerate}
\item So far, we analyzed usage regularities of threshold concepts in such a way that we assumed a one-to-one mapping between selected words and the corresponding concepts: 
for example, the lemma /\textit{cost}/ then stands directly and uniquely for the corresponding concept of cost.
This is where we can start and develop a more general two-step procedure that assumes that concepts can be lexically named by groups of words that form a sort of paradigm of lexical paraphrases of the same concept.
This view locates lexical naming alternatives for concepts above the level of lexeme groups but below the level of word fields.
Using the apparatus of word embeddings, such lexeme clusters can be computed as cliques of words with very high cosine similarities of their embeddings, that is, clusters of paradigmatically strongly associated words. 
However, one should not underestimate the amount of post-correction required to clean up such clusters, for example to sort out highly associated words that do not designate the concept underlying the cluster.
In any case, such a procedure makes it possible to identify further co-texts within which the same threshold concept is specified. This would mean to considerably enlarge the database of threshold concept research. 
Ideally, this approach would also include non-lexical paraphrases.
\item A second extension concerns the detailed consideration of \emph{basic}-, \emph{discipline}- and \emph{procedural}-level concepts.
More specifically, formal language corpora should be divided into subsets of texts, which are either at the basic, disciplinary or procedural level.
In this way, we gain access to contexts of use of threshold concepts that allow us to assign them to one of these levels or to determine linguistic evidence of what was described above as conceptual change, that is, the transition in the use of a concept between these levels that might indicate a higher dynamics relevant to formal learning contexts.
\item A third extension concerns the broadening of the basis of comparison of threshold concepts.
That is, instead of just networking them with each other, we could additionally examine how they network with non-threshold concepts or with concepts that belong to one of the three basic, disciplinary or procedural levels.
In any event, this should again be done in such a way that each of these reference sets is small and selected in advance in order to allow transparent comparisons.
%

\end{enumerate}
\color{black}

\section{General discussion}
\label{sec:general-discussion}

Different resources make different claims about threshold concepts.
For the 63 threshold concept expressions $t_1, \ldots, t_{63}$ under consideration, this claim can be represented in the form:  \enquote{sense($t_1$) is related to sense($t_2$), sense($t_2$) is more related to sense($t_8$)}, and so on, where the degree of relatedness differs between the corpora (cf. Subsec.~\ref{subsubsec:dictionary-concepts}).
That is, different resources express a different \enquote{take on threshold concepts}.
This in turn leads to the question whether the different resources also lead to a (crucially) different understanding of threshold concepts for learners. 
The computational linguistic assessment therefore has implications for text comprehension \citep{Kintsch:1988} and domain learning \citep{Alexander:2018}.
The results are also related to findings on learners' understanding of the threshold concepts \citep{Brueckner:Zlatkin-Troitschanskaia:2018}. 
However, there is no straightforward mapping between dictionary concepts and indexed concepts (a student's private understanding of lexical meanings). 
%

\citet{Oakhill:Cain:Bryant:2010} show in a study on language development that word reading and text comprehension are dissociated.
This implies that text comprehension and word decoding follow different developmental trajectories and can be taught at least to some degree independently. 
The acquisition of threshold concepts proceeds at least on these two routes, meaning that developing respective understandings draws on text comprehension as well as on lexical definitions. 
This line of thought emphasizes the need for a semantic analysis of threshold concepts in business education (which, as far as we know, is missing, see above), either in form of componential analyses or paraphrases/definitions.

Text comprehension is not only based on memory processes but also on constructionist processes \citep{vanDenBroek:Rapp:Kendeou:2005}. 
The latter can, for instance, arise due to associations bound up with readers' indexed concepts.
This includes personal preferences as well as all sorts of top-down processes. 
Constructionist aspects of comprehension are bound up with learners' everyday language and prior knowledge. 
Liken the acquisition of a specialized language to second language acquisition (for such a view see  Subsec.~\ref{subsec:threshold-and-special-vocab}), this implies that also the first language should be taken into account.
Here we meet advice from educational research, namely \enquote{Alltags- und Fachsprache als je für sich entwicklungsfähig anzusehen und im Unterricht zu thematisieren} (\textit{to regard everyday and specialized language as being capable of development in their own right and to address them both in class} [translated by AL]) \citep[235]{Rincke:2010}.

The prior knowledge of learners also plays a role in reading hypertexts such as Wikipedia articles. 
Interestingly, using hypertexts as a learning resource can be advantageous in particular for informed learners, since the hypertext structure allows them to exert a strategic reading processes \citep{Salmeron:Kintsch:Canas:2006}.
That is, online (hypertext) resources such as Wikipedia can enrich the learning landscape for education (as they already do as a matter of fact, cf. Subsec.~\ref{subsec:formal-informal}).

\section{Conclusion}
\label{sec:conclusion}

%
%
The computational linguistic perspective adopted in the present contribution pursues an orientation which, in terms of educational research on threshold concepts, has two special features. On the one hand, it complements content analyses, which are classically used to analyze textbooks, protocols or other textually and graphically represented materials in order to work out education-related meanings from the materials \citep[e.g.][]{Krippendorff:2013}. 
%
The often tedious and lengthy manual evaluation with only a limited number of documents and the corresponding susceptibility to errors is as a matter of fact limited to a small amount of data. 
Computational linguistic analyses, to the contrary, can process huge corpora. 
Secondly, so-called \emph{utilization-of-learning-opportunities} models are used to model the mechanisms of action of teaching-learning arrangements in educational research \citep[e.g][]{Braun:Weiss:Seidel:2014}. These models show the interactions between learning-relevant aspects in terms of input-process-output paths. Very often learning outcomes are analyzed in connection with different input factors (e.g. socio-economic status, gender, intelligence, self-assessed use of learning media). Significantly less frequently, however, the learning potentials of the respective learning environments or learning materials are considered independently of a learner's assessment. With the computational linguistic approach presented here, especially the learning media that are used as input into the learning processes are processed on a large scale and thus a description of the learning environment is presented that can be considered in informal as well as formal learning processes.
Ultimately learning, the meaning of threshold concept expressions and their use in text resources are embraced within the contour of an emerging research program -- encompassing specialized vocabularies, learning and education, and computational linguistics -- in terms of mental, referential and differential meanings. The latter two (referential and differential meanings) are used in order to derive hypotheses concerning formal and informal learning contexts with respect to a special class of expressions, viz. threshold concepts.
A second focus was the development of a computational linguistic model for operationalizing threshold concepts for the analysis of learning resources. In this context, we developed the notion of a Threshold Concept Network (TCN) and quantified it by means of alpha-cuts, taking into account the  \enquote{web of a threshold concepts} \citep{Davies:Mangan:2007}.
In this way, we were able to prove an exceptional status of threshold concepts in textbooks, at least at the node level. 
The main result was that formal and informal resources can indeed be distinguished in terms of their threshold concepts' profiles.
Furthermore, Wikipedia turns out to be a first class formal learning resource.
Continuing this line of research will include at least the following steps: the methodological considerations discussed in Subsec.~\ref{subsec:discussion} are to be addressed. A lexical semantic analysis of threshold concepts is due. And, most importantly, our findings have to be tied back to education assessments of learners. 
Furthermore, experimental studies have to be designed that investigate systematically the impact of different resources on learning. 
%
Very often experimental studies are developed on assumptions that have not been tested themselves. On the basis of the computational linguistic assessment, however, it is possible to develop more specific questions.
Most notably, the threshold concept acquisition of learners  can be compared depending on the media to learn (e.g. Wikipedia vs textbook vs daily newspaper, and their interaction and complementary uses) -- whereby, of course, the corresponding media competencies and information literacy or other (intellectual) characteristics must also be controlled \citep{Vernooij:2000}. 
The assessments from the study presented here provide a starting point for such experiments which in turn would round out the emerging research program we sketched. 

\section*{Conflict of Interest Statement}

The authors declare that the research was conducted in the absence of any commercial or financial relationships that could be construed as a potential conflict of interest.

\section*{Author Contributions}


AL mainly has written Secs.~\ref{sec:rationale-method} and \ref{sec:general-discussion}, and performed the  Kolmogorov-Smirnov goodness-of-fit tests and the power law fitting (Tabs.~\ref{tab:ks-all},
\ref{tab:ks-only},
\ref{tab:ks-without},
\ref{tab:Power-law-fitting},
\ref{tab:cohesion-x},
\ref{tab:cws-x}, 
\ref{tab:cohesion-cbow-y}
) 
SB selected the textbook corpus and mainly has written  Subsecs.~\ref{subsec:conceptual-change} and \ref{subsec:threshold-business-economics}.
SB and AM have jointly written Subsec.~\ref{subsec:troublesome-language} (AL just added three references).
AL and SB have jointly written Subsec.~\ref{subsec:formal-informal}.
GA carried out the preprocessing of the corpora and the word embeddings.
TU calculated the compound distributions and the threshold concept networks.
AM designed the computational linguistic measurement procedure for threshold concepts, implemented the corresponding network analyses, has written almost all parts of Sec.~\ref{sec:study} and generated Figures 
\ref{fig:MinEdgeWeightVsCohesion}, \ref{fig:MinEdgeWeightVsCWS} and \ref{fig:MinEdgeWeightVsCohesionCBOW}.
Sec.~\ref{sec:conclusion} has been jointly written by SB, AL and AM.

\section*{Funding}
Details of all funding sources should be provided, including grant numbers if applicable. Please ensure to add all necessary funding information, as after publication this is no longer possible.



\section*{Data Availability Statement}

Due to copyright restrictions, the newspaper and the textbook corpora are not publicly available. Wikipedia can be obtained via Wikipedia dumps.


\section*{Appendix A: Threshold concepts used in the study}

From 
\citet{Hoadley:Tickle:Wood:Kyng:2015}, \citet{Davies:Mangan:2007}, 
\citet{Sender:2017}, 
\citet{Lucas:Mladenovic:2009}, and \citet{Brueckner:Zlatkin-Troitschanskaia:2018} we compiled the following list of threshold concept expressions:
                    \enquote{Einkommen},
                     \enquote{Bargeld},
                     \enquote{Gewinn},
                     \enquote{investieren},
                     \enquote{Investition},
                     \enquote{sparen},
                     \enquote{Abschreibung},
                     \enquote{Intertemporalität},
                     \enquote{intertemporal},
                     \enquote{Kumulationseffekt},
                     \enquote{Opportunitätskosten},
                     \enquote{Marginalanalyse},
                     \enquote{Elastizität},
                     \enquote{Marktgleichgewicht},
                     \enquote{Marktinteraktion},
                     \enquote{Wohlfahrt},
                     \enquote{Effizienz},
                     \enquote{Nachhaltigkeit},
                     \enquote{Aufwendung},
                     \enquote{Auszahlung},
                     \enquote{Kosten},
                     \enquote{Informationsasymmetrie},
                     \enquote{Leverage},
                     \enquote{Marktstruktur},
                     \enquote{Preisgestaltung},
                     \enquote{Risikoaversion},
                     \enquote{Rendite},
                     \enquote{Trade-offs},
                     \enquote{Arbitrage},
                     \enquote{Cashflow},
                     \enquote{Diversifikation},
                     \enquote{Hedging},
                     \enquote{Sicherungsgeschäft},
                     \enquote{Markteffizienz},
                     \enquote{Risiko},
                     \enquote{Zeitwert},
                     \enquote{Nutzen},
                     \enquote{Risikopräferenz},
                     \enquote{Preisgestaltungsmodell},
                     \enquote{Preisbildungsmodell},
                     \enquote{Liquidität},
                     \enquote{Nominal},
                     \enquote{real},
                     \enquote{Derivative},
                     \enquote{Principial-Agent},
                     \enquote{Grenzkosten},
                     \enquote{Marginalkosten},
                     \enquote{Break-Even},
                     \enquote{Produktlebenszyklus},
                     \enquote{Diamantenmodell},
                     \enquote{SWOT},
                     \enquote{Optimierung},
                     \enquote{Brutto},
                     \enquote{brutto},
                     \enquote{Netto},
                     \enquote{netto},
                     \enquote{Umsatzerlös},
                     \enquote{Ertrag},
                     \enquote{Erlös},
                     \enquote{Bruttoinlandsprodukt},
                     \enquote{BIP},
                     \enquote{marginal},
                     \enquote{Erwartungswert}.
All these threshold concepts have been used in the study.

\section*{Appendix B: Textbook corpus}
\begin{enumerate}
\item Blum, Ulrich (2017). \textit{Grundlagen der Volkswirtschaftslehre}. Berlin and Boston: De Gruyter Oldenbourg.

\item Breyer, Friedrich (2020). \textit{Mikroökonomik. Eine Einführung}. 7., revised and updated ed. Berlin and Heidelberg: Springer Gabler. doi: 10.1007/978-3-662-60779-4.

\item Buchholz, Ulrike and Susanne Knorre (2019). \textit{Interne Kommunikation und Unternehmensführung. Theorie und Praxis eines kommunikationszentrierten Managements}. Wiesbaden: Springer Gabler. doi: 10.1007/978-3-658-23432-4.

\item Meffert, Heribert et al. (2019). \textit{Marketing: Grundlagen marktorientierter Unternehmensführung}. Konzepte -- Instrumente -- Praxisbeispiele. Wiesbaden: Springer Gabler. doi: 10.1007/978-3-658-21196-7.

\item Mumm, Mirja (2015). \textit{Kosten- und Leistungsrechnung. Internes Rechnungswesen für Industrie- und Handelsbetriebe}. 2., revised and expanded. Berlin and Heidelberg: Springer Gabler. doi: 10.1007/978-3-662-44379-8.

\item Neubäumer, Renate, Brigitte Hewel, and Thomas Lenk, eds. (2017). \textit{Volkswirtschaftslehre. Grundlagen der Volkswirtschaftstheorie und Volkswirtschaftspolitik}. 6th ed. Springer Gabler. doi: 10.1007/978-3-658-16523-9.

\item Pfannmöller, Jürgen (2018). \textit{Kreative Volkswirtschaftslehre. Eine handlungs- und praxisorientierte Einführung in die Volkswirtschaftslehre}. Wiesbaden: Springer Gabler. doi: 10.1007/978-3-658-07958-1.

\item Schierenbeck, Henner and Claudia B. Wöhle (2016). \textit{Grundzüge der Betriebswirtschaftslehre}. 19., revised. München: De Gruyter Oldenbourg.

\item Sellenthin, Mark (2017). \textit{Volkswirtschaftslehre -- mathematisch gedacht}. Wiesbaden: Springer Gabler. doi: 10.1007/978-3-658-13905-6.

\item Thielscher, Christian (2014). \textit{Wirtschaftswissenschaften verstehen. Eine Einführung in ökonomisches Denken}. Wiesbaden: Springer Gabler.

\item Thommen, Jean-Paul et al. (2017). \textit{Allgemeine Betriebswirtschaftslehre. Umfassende Einführung aus managementorientierter Sicht}. 8., completely revised ed. Wiesbaden: Springer Gabler.

\item Varian, Hal R. (2016). \textit{Grundzüge der Mikroökonomik}. 9., updated and expanded ed. Berlin and Boston: Walter de Gruyter.

\item Weeber, Joachim (2015). \textit{Einführung in die Volkswirtschaftslehre: Für den Bachelor}. 3., updated and expanded ed. Berlin and Boston: Walter de Gruyter.

\item Woeckener, Bernd (2019). \textit{Volkswirtschaftslehre. Eine Einführung}. 3., revised and expanded ed. Berlin and Heidelberg: Springer Gabler. doi: 10.1007/978-3-662-59222-9.
\end{enumerate}

\bibliography{threshold.bib}

\end{document}